\documentclass[]{oppo}
\usepackage{float}

\usepackage{times}
\usepackage{latexsym}
\usepackage{listings}
\usepackage[T1]{fontenc}
\usepackage[utf8]{inputenc}

\usepackage{microtype}

\usepackage{inconsolata}

\usepackage{graphicx}  
\usepackage{booktabs}  
\usepackage{multirow}  
\usepackage{array}     
\usepackage{siunitx}   
\usepackage[utf8]{inputenc} 
\usepackage[T1]{fontenc}    
\usepackage{hyperref}       
\usepackage{url}            
\usepackage{booktabs}       
\usepackage{amsfonts}       
\usepackage{nicefrac}       
\usepackage{microtype}      
\usepackage{xcolor}         
\usepackage{xspace}
\usepackage{fancyhdr}
\usepackage{ifthen}
\usepackage{wrapfig}

\usepackage{etoolbox}
\makeatletter
\patchcmd{\@maketitle}
  {\@title}
  {\@title\\[-0.25ex]\normalsize\bfseries \@author}
  {}{}
\makeatother

\usepackage{times}
\usepackage{latexsym}
\usepackage{minted}
\usepackage{placeins} 
\usepackage[T1]{fontenc}
\usepackage[utf8]{inputenc}
\usepackage{microtype}
\usepackage{inconsolata}
\usepackage{graphicx}
\usepackage{wrapfig}
\usepackage{fontawesome5}
\usepackage{multirow}
\usepackage{booktabs}
\usepackage{makecell}
\usepackage{enumitem}
\usepackage[table,xcdraw]{xcolor}
\usepackage{colortbl}
\usepackage{amsmath,amssymb} 
\usepackage{lipsum} 
\usepackage{array}
\usepackage{longtable,tabularx}
\usepackage{tcolorbox}
\tcbuselibrary{skins,breakable}
\usepackage{algorithm}
\usepackage{algpseudocode}
\usepackage{listings}
\usepackage{float}
\usepackage{multicol}
\usepackage{siunitx}   
\usepackage{fancyvrb}
\usepackage{hyperref}
\usepackage{xcolor}
\newcommand{\best}[1]{{\color{red}#1}}
\newcommand{\second}[1]{{\color{blue}#1}}

\providecommand{\inst}[1]{}

\newcommand{\taxonomy}[1]{\textsc{DEFT}}
\newcommand{\bench}[1]{\textsc{Finder}}

\usepackage{tikz}
\usepackage{pgfplots}
\pgfplotsset{compat=1.18}

\setlist[itemize]{topsep=0pt, partopsep=0pt, parsep=0pt, itemsep=1pt, leftmargin=*}

\title{UniCSG: Unified High-Fidelity Content-Constrained Style-Driven Generation via Staged Semantic and Frequency Disentanglement}

\author[1]{Jingwei Yang}
\author[2]{Ruoxi Wu}
\author[2]{Wei Shen}
\author[2]{Meng Li}
\author[2]{Yulong Liu}
\author[2]{Huimin She}
\author[2]{Lunxi Yuan}

\affiliation[1]{China University of Mining and Technology, Beijing, China}
\affiliation[2]{OPPO Artificial Intelligence Center, Beijing, China}

\abstract{
Style transfer must match a target style while preserving content semantics. DiT-based diffusion models often suffer from content–style entanglement, leading to reference-content leakage and unstable generation. We present UniCSG, a unified framework for content-constrained, style-driven generation in both text-guided and reference-guided settings. UniCSG employs staged training: (i) a latent-space semantic disentanglement stage that combines low-frequency preprocessing with conditioning corruption to encourage content–style separation, and (ii) a latent-space frequency-aware detail reconstruction stage that refines details via multi-scale frequency supervision. We further incorporate pixel-space reward learning to align latent objectives with perceptual quality after decoding. Experiments demonstrate improved content faithfulness, style alignment, and robustness in both settings.

\vspace{10pt}
\textbf{Code \& Data \faGithub}: 
\url{https://github.com/Jankinwei/UniCSG}

\textbf{Correspondence}: Ruoxi Wu at \url{wuruoxi@oppo.com}}

\begin{document}
\maketitle

\section{Introduction}
\label{sec:intro}

Style transfer\cite{Gatys_2016_CVPR} is a core controllable generation task. Text-guided style transfer applies a language-specified style to a content image, whereas reference-guided style transfer transfers the visual appearance of a reference image to the target content. In applications such as art creation, advertising, and game development, the goal is to preserve content semantics while achieving high-fidelity style alignment\cite{zhang2026qwenstylecontentpreservingstyletransfer}.

Most diffusion-based style transfer systems rely on either SDXL-style UNets\cite{ICLR2024_081b0806} or DiT backbones\cite{esser2024scaling,flux2024,Peebles_2023_ICCV}. UNets benefit from local inductive biases\cite{pmlr-v285-zhang24a,zhang2024forgedit} and often yield better empirical content–style separation\cite{wang_styleadapter_2025,zhang2025cdstcolordisentangledstyle}, yet their scalability and global modeling capacity can limit performance at high resolution and in complex scenes. DiT architectures scale better, but they typically lack fine-grained disentanglement mechanisms\cite{zhang2026qwenstylecontentpreservingstyletransfer}. As a result, they can suffer from style overfitting\cite{Lei_2025_CVPR}, reference-content leakage\cite{xing2025csgo,wu2025uso,chen2025consislora}, and instability such as checkerboard artifacts\cite{Lei_2025_CVPR,zhang2025ustyditultrahighqualityartistic,Li_2025_ICCV}.

We attribute these issues to three factors. First, latent diffusion models are often trained with a single reconstruction objective, which biases learning toward low-level textures and under-emphasizes high-level semantics such as contours and layout\cite{vtp}. Second, frequency coupling further entangles content and style, reducing controllability and fidelity. DiT backbones tend to model low-frequency structure more effectively than high-frequency textures\cite{ma2025decofrequencydecoupledpixeldiffusion}. Third, optimization occurs in latent space whereas evaluation is performed in pixel space. Latent-optimal solutions can diverge from perceptual optimality after decoding.

\begin{center}
\includegraphics[width=\textwidth]{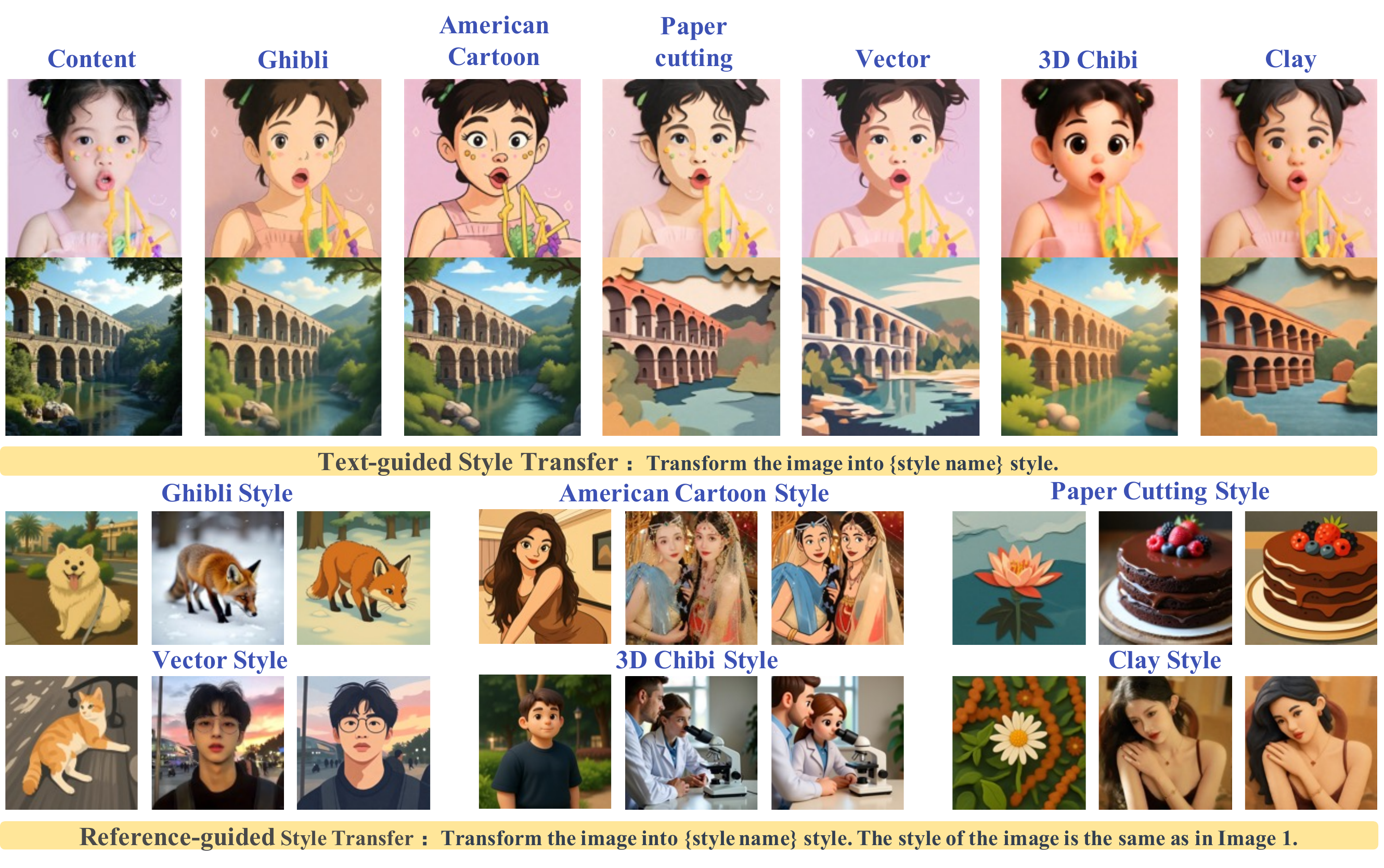}
\captionof{figure}{Teaser of UniCSG on text-guided and reference-guided style transfer. Given a content image, UniCSG transforms it into user-specified styles under both text prompts and reference exemplars, showing faithful content preservation and style alignment. In the reference-guided rows, each triplet shows (left to right) the reference image, the content image, and the generated result.}
\label{fig:teaser}
\end{center}

To address these challenges, we propose UniCSG, a unified framework for high-fidelity, content-constrained, style-driven generation. UniCSG follows a staged, progressive paradigm. In latent space, two sequential stages with decoupled objectives handle semantic disentanglement and frequency-aware detail reconstruction, respectively. A pixel-space perceptual alignment module further bridges the latent–pixel objective gap.

Latent-space Stage 1 (Semantic Disentanglement): we combine low-frequency preprocessing with conditioning corruption to encourage the model to rely on high-level cues, improving content–style disentanglement and reducing content leakage.

Latent-space Stage 2 (Frequency-aware Detail Reconstruction): we introduce multi-scale frequency decomposition and a frequency-weighted objective to refine high-frequency textures.

Pixel-space Reward Learning (Reward-guided Perceptual Alignment): we incorporate pixel-space reward learning to improve perceptual quality after decoding.

Through this disentangle–reconstruct–align pipeline, UniCSG jointly improves controllability, detail fidelity, and perceptual quality.

Our contributions can be summarized as follows:

\begin{itemize}
\item We introduce a \textbf{semantic disentanglement} strategy based on low-frequency preprocessing and conditioning corruption, which effectively separates style from content and prevents reference-content leakage.

\item We propose a \textbf{frequency-aware detail reconstruction} mechanism that leverages multi-scale frequency decomposition and frequency-weighted supervision to enhance high-frequency texture recovery upon the semantic scaffold.

\item We introduce a \textbf{reward-guided learning} mechanism in pixel space that bridges the gap between latent-space optimization objectives and perceptual quality after decoding.

\end{itemize}

\section{Related Works}
\subsection{Content-Constrained Style-Driven Generation}
Early CNN-based style transfer defines separate content and style losses to constrain structural and textural similarity, respectively\cite{Gatys_2016_CVPR}. Later methods add attention\cite{Deng_2022_CVPR,Shang_2025_CVPR} and multi-scale fusion\cite{huang2017adain} for finer local control. With diffusion models, the focus shifts to imposing precise content constraints on powerful backbones. IP-Adapter~\cite{ye2023ip-adapter} injects reference features via cross-attention for disentangled control. ControlNet~\cite{Zhang_2023_ICCV} conditions on spatial cues (edges, depth) to preserve structure. However, most approaches still operate at the pixel or shallow-feature level without explicit high-level semantic constraints, causing structural distortion and content-style entanglement in complex scenes. Frequency coupling further limits fidelity and visual quality. These limitations motivate stronger semantic constraints and frequency-aware optimization for robust, high-fidelity style transfer.

\subsection{Semantic Disentanglement}
In image generation, pixel-level reconstruction alone often drives latents to overfit textures and noise while under-capturing high-level semantics\cite{vtp}. Yet latent semantic capacity correlates strongly with generation quality. Prior work addresses this gap by strengthening encoders (e.g., VAE–ViT hybrids or unified encoders\cite{vtp}), aligning latents with semantic features via DINOv2\cite{oquab2024dinov}, or employing asynchronous denoising such as SFD\cite{Pan2025SFD}. In contrast, we retain the standard encoder and decouple objectives via staged training. Stage 1 learns semantic disentanglement under degraded inputs. Stage 2 refines details guided by the learned scaffold.

\subsection{Frequency Disentanglement}
Diffusion Transformers have popularized high/low-frequency separation. DeCo\cite{ma2025decofrequencydecoupledpixeldiffusion}, DiP\cite{chen2025diptamingdiffusionmodels}, PixelDiT\cite{yu2025pixelditpixeldiffusiontransformers}, and JiT\cite{li2026basicsletdenoisinggenerative} assign different components to low-frequency semantics versus high-frequency textures, aligning with Transformers' strengths in low-frequency modeling. Common strategies employ heavy downsampling\cite{ma2025decofrequencydecoupledpixeldiffusion,yu2025pixelditpixeldiffusiontransformers} or embedding bottlenecks\cite{li2026basicsletdenoisinggenerative} to constrain the backbone to low frequencies. Lightweight decoders\cite{ma2025decofrequencydecoupledpixeldiffusion} or auxiliary paths\cite{yu2025pixelditpixeldiffusiontransformers} then restore high-frequency textures. Another line decouples frequencies in latent space\cite{Wang_2025_ICCV,xiang2025denoisingvisiontransformerautoencoder}, exploiting early denoising for low frequencies and later steps for high frequencies\cite{yin2025ferafrequencyenergyconstrainedrouting}. We deepen this perspective with a two-stage paradigm. Stage 1 concentrates on low-frequency structure and content–style integration. Stage 2 targets frequency-aware detail reconstruction, yielding stronger disentanglement and higher fidelity.
\section{Methodology}

\begin{figure*}[!t]
\centering
\includegraphics[width=\textwidth]{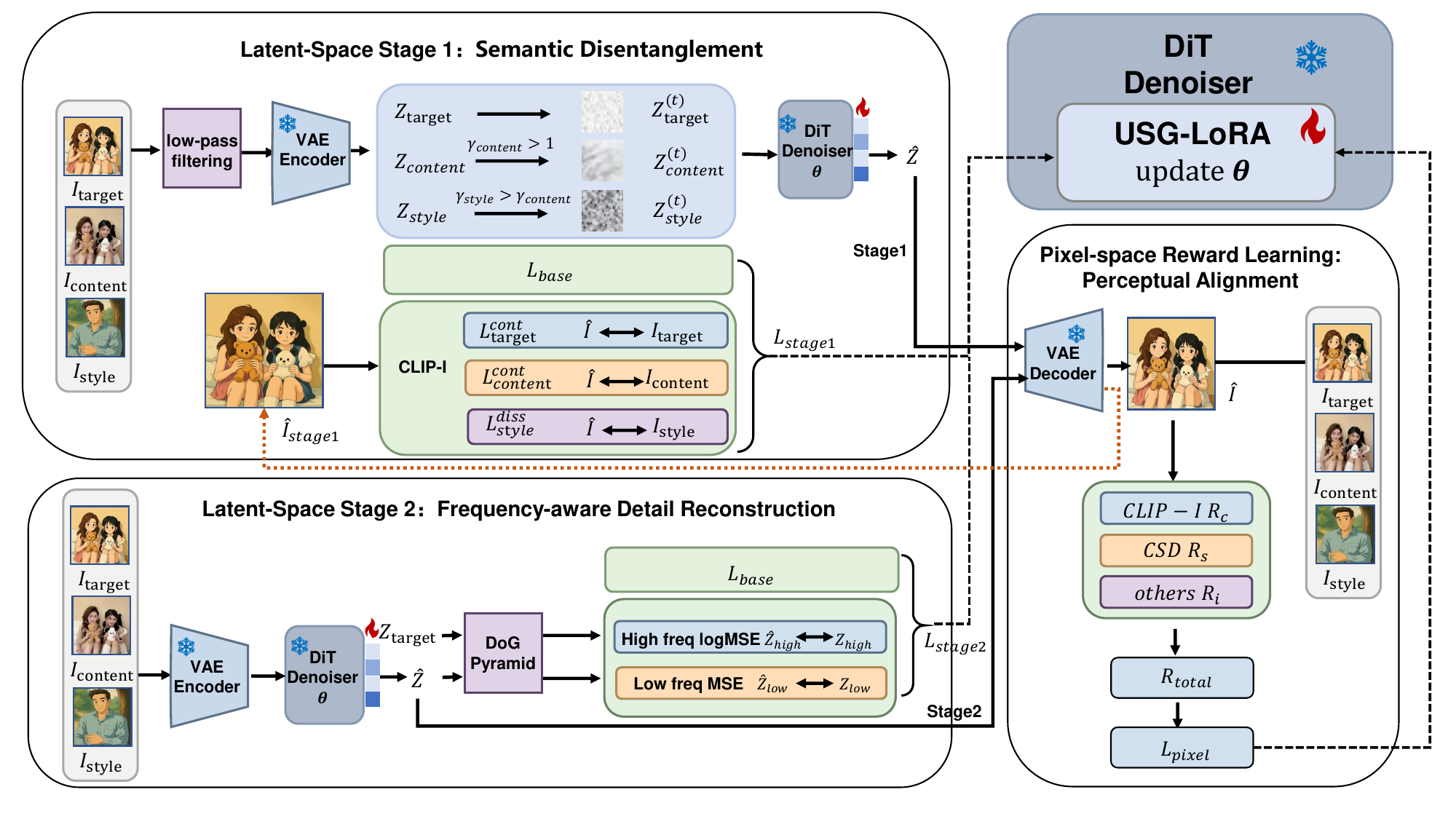}
\caption{Overview of UniCSG. We use staged semantic--frequency disentanglement in latent space and reward-guided perceptual alignment in pixel space. Stage 1 (semantic disentanglement): low-frequency preprocessing plus conditioning corruption builds a robust content--style scaffold. Stage 2 (frequency-aware detail reconstruction): multi-scale frequency supervision refines high-frequency textures atop the scaffold. Pixel-space reward learning further matches latent objectives to perceptual quality after decoding.}
\label{fig:main}
\end{figure*}

We build UniCSG on top of a pretrained image-editing model (Qwen-Image-Edit-2509\cite{wu2025qwenimagetechnicalreport}). We unify text-to-image and image-to-image settings with a four-tuple input $\langle text, ref\_img, source\_img, gt\_img \rangle$. Both text-guided and reference-guided stylization follow the same content-constrained generation framework, differing only in the source of style information: text-guided transfer uses the text prompt to specify the desired style, whereas reference-guided transfer uses the reference image as the style exemplar. We structure the methodology description into three core sections. Semantic Disentanglement (Stage 1) is covered in Sec.~\ref{sec:Semantic}. Sec.~\ref{sec:Frequency} addresses Frequency-aware Detail Reconstruction (Stage 2). Pixel-space Reward Learning is then presented in Sec.~\ref{sec:reward}. An overview is shown in Fig.~\ref{fig:main}.

\subsection{Semantic Disentanglement}
\label{sec:Semantic}
This stage targets semantic entanglement between content and style. We combine low-frequency preprocessing with conditioning corruption to increase the difficulty of the conditioning inputs. This encourages the model to learn disentangled content–style representations and a stable semantic scaffold that reduces content leakage.

Diffusion denoising naturally follows a coarse-to-fine trajectory: early steps primarily recover global contours and semantic layout\cite{yin2025ferafrequencyenergyconstrainedrouting}. Accordingly, we apply low-pass filtering\cite{lai2025diffusiblehighdimensionallatentspaces,lee2025frequencyawaretokenreductionefficient} to the input images before VAE encoding, retaining only frequencies below a threshold $\tau$ (default $\tau{=}0.2$) for the target $\mathbf{I}_{\text{target}}$ and the conditioning images $\mathbf{I}_{\text{content}}$ and $\mathbf{I}_{\text{style}}$:

\begin{equation}
\begin{aligned}
\mathbf{I}^{\text{low}}_{\text{target}}&=\mathcal{F}_{\text{low\_pass}}(\mathbf{I}_{\text{target}}),\\
\mathbf{I}^{\text{low}}_{\text{content}}&=\mathcal{F}_{\text{low\_pass}}(\mathbf{I}_{\text{content}}),\\
\mathbf{I}^{\text{low}}_{\text{style}}&=\mathcal{F}_{\text{low\_pass}}(\mathbf{I}_{\text{style}}).
\end{aligned}
\label{eq:lowpass}
\end{equation}
This operation encourages the model to focus on low-frequency structure early in training, supporting stable semantic learning before high-frequency refinement\cite{Gao_2025_ICCV}.

To achieve precise content–style disentanglement, we introduce conditioning corruption under a unified timestep. Specifically, we keep a single timestep condition per training step and apply different corruption strengths to the content and style conditionings. This yields an information hierarchy: the target is the least corrupted, content is moderately corrupted, and style is the most corrupted.

To remain fully compatible with the pretrained noise schedule, we sample a unified timestep $t$ at each training step and use the scheduler-provided noise coefficient $\sigma(t)\in[0,1]$ (denoted $\sigma_t$). For clarity, we denote the clean target/content/style latents by $\mathbf{z}_{\text{target}}$, $\mathbf{z}_{\text{content}}$, and $\mathbf{z}_{\text{style}}$, and use the superscript $(t)$ to indicate the noised latent at timestep $t$. Under this unified timestep, the target latent follows the standard convex-combination noising process:

\begin{equation}
\mathbf{z}^{(t)}_{\text{target}} = (1-\sigma_t)\mathbf{z}_{\text{target}} + \sigma_t \boldsymbol{\epsilon}_{\text{target}},\quad \boldsymbol{\epsilon}_{\text{target}}\sim \mathcal{N}(0,\mathbf{I}).
\label{eq:ztar}
\end{equation}
Rather than using different timesteps, we express the desired information hierarchy by applying stronger corruption to the conditioning branches under the same timestep. Specifically, we introduce noise amplification factors $\gamma_i$ for $i \in \{\text{content}, \text{style}\}$, where $\gamma_{\text{style}} > \gamma_{\text{content}} > 1$. These factors are implemented via higher-variance Gaussian noise. For a given amplification factor $\gamma_i$, we construct the amplified noise as:

\begin{equation}
\begin{aligned}
\boldsymbol{\epsilon}_{\gamma_i}&=\boldsymbol{\epsilon}+\sqrt{\gamma_i^2-1}\,\boldsymbol{\epsilon}',\quad 
\boldsymbol{\epsilon},\boldsymbol{\epsilon}'\sim\mathcal{N}(0,\mathbf{I}),\ \text{independent}\\
&\Rightarrow\ 
\boldsymbol{\epsilon}_{\gamma_i}\sim\mathcal{N}(0,\gamma_i^2\mathbf{I}).
\end{aligned}
\label{eq:noise_amp}
\end{equation}
Under the same timestep $t$, the noised content and style conditionings are defined as:

\begin{equation}
\begin{aligned}
\mathbf{z}^{(t)}_{\text{content}} &= (1-\sigma_t)\mathbf{z}_{\text{content}} + \sigma_t \boldsymbol{\epsilon}_{\gamma_{\text{content}}},\\
\mathbf{z}^{(t)}_{\text{style}} &= (1-\sigma_t)\mathbf{z}_{\text{style}} + \sigma_t \boldsymbol{\epsilon}_{\gamma_{\text{style}}}.
\end{aligned}
\label{eq:zcon_zsty}
\end{equation}
Consequently, under the same timestep $t$, the style conditioning is more corrupted than the content conditioning, while the target branch remains the least corrupted. This $\gamma_i$-controlled information hierarchy encourages the model to extract global style cues from heavily corrupted style inputs and structural semantics from less corrupted content inputs, suppressing reference-content leakage within a single-timestep training framework\cite{chefer2026selfsupervisedflowmatchingscalable}.

To improve generalization, we introduce a probabilistic replacement mechanism: with probability $p$, we replace the content or style conditioning with pure noise, forcing the model to generate plausible structure under partial information loss. To smoothly transition from Stage 1 disentanglement learning to Stage 2 full-detail reconstruction, we apply a progressive decay schedule to the corruption strength during training.

The objective in this stage combines a base reconstruction loss $\mathcal{L}_{\text{base}}$ with a semantic loss $\mathcal{L}_{\text{sem}}$ that guides content–style disentanglement:

\begin{equation}
\mathcal{L}_{\text{stage1}} = \mathcal{L}_{\text{base}} + \mathcal{L}_{\text{sem}}.
\label{eq:stage1_overview}
\end{equation}
The base loss is the standard velocity-prediction objective:

\begin{equation}
\mathcal{L}_{\text{base}} = w(t)\cdot \mathbb{E}\Big[\big\| \mathbf{v}_\theta(\mathbf{z}^{(t)}_{\text{target}},t,\mathbf{z}^{(t)}_{\text{content}},\mathbf{z}^{(t)}_{\text{style}}) - (\boldsymbol{\epsilon}_{\text{target}}-\mathbf{z}_{\text{target}}) \big\|_2^2\Big].
\label{eq:base_loss}
\end{equation}
The semantic loss uses CLIP\cite{pmlr-v139-radford21a} image features $\phi(\cdot)$ to define three complementary signals. The target-faithfulness term $\mathcal{L}^{\text{cont}}_{\text{target}}$ aligns the generated output with the ground-truth target at a semantic level. The content-preservation term $\mathcal{L}^{\text{cont}}_{\text{content}}$ enforces the core objects and scene layout of the content image. The content-style repulsion term $\mathcal{L}^{\text{diss}}_{\text{style}}$ reduces semantic similarity with the style image, mitigating unintended content transfer from the style reference:

\begin{align}
\mathcal{L}_{\text{sem}} =\;&
  \lambda^{\text{cont}}_{\text{target}} \cdot \underbrace{\mathbb{E}\Big[1 - \text{sim}_{\text{CLIP-I}}\big(\phi(\hat{\mathbf{I}}),\, \phi(\mathbf{I}_{\text{target}})\big)\Big]}_{\mathcal{L}^{\text{cont}}_{\text{target}}} \nonumber\\
&+ \lambda^{\text{cont}}_{\text{content}} \cdot \underbrace{\mathbb{E}\Big[1 - \text{sim}_{\text{CLIP-I}}\big(\phi(\hat{\mathbf{I}}),\, \phi(\mathbf{I}_{\text{content}})\big)\Big]}_{\mathcal{L}^{\text{cont}}_{\text{content}}} \nonumber\\
&+ \lambda^{\text{diss}}_{\text{style}} \cdot \underbrace{\mathbb{E}\Big[\text{sim}_{\text{CLIP-I}}\big(\phi(\hat{\mathbf{I}}),\, \phi(\mathbf{I}_{\text{style}})\big)\Big]}_{\mathcal{L}^{\text{diss}}_{\text{style}}}.
\label{eq:sem_loss}
\end{align}

In summary, conditioning corruption and semantic supervision promote content–style disentanglement and provide a semantic foundation for Stage 2.

\subsection{Frequency-aware Detail Reconstruction}
\label{sec:Frequency}
This stage focuses on high-fidelity reconstruction of fine-grained textures after the model has learned a stable semantic blueprint. We guide detail synthesis through multi-scale frequency decomposition and a carefully designed objective emphasizing high-frequency refinement.

Once the model has learned the basic structure, we remove the low-pass constraint and train with full-spectrum information. The key is to introduce frequency-domain losses as explicit guidance. Since the denoising objective is defined in latent space, we perform frequency decomposition directly on latents for consistency with the training target. Specifically, we apply a DoG (Difference of Gaussian) pyramid to decompose the predicted latent $\hat{\mathbf{z}}$ and the target latent $\mathbf{z}_{\text{target}}$ into multi-scale components\cite{Wang_2025_ICCV,xiang2025denoisingvisiontransformerautoencoder}. This separates low-frequency parts $\hat{\mathbf{z}}^{(k)}_{\text{low}}$, $\mathbf{z}^{(k)}_{\text{target, low}}$ and high-frequency parts $\hat{\mathbf{z}}^{(k)}_{\text{high}}$, $\mathbf{z}^{(k)}_{\text{target, high}}$, where $k$ denotes the scale level. Any perceptual gap introduced by operating in latent space is subsequently addressed by the pixel-space reward learning (Sec.~\ref{sec:reward}).

Low-frequency components correspond to transferable global structure with relatively uniform error distributions. We therefore use MSE to provide strong gradients for accurate global reconstruction. In contrast, high-frequency components are inherently mixed: they include both stylistic fine textures to be transferred and irrelevant noise or incidental details. For high-frequency terms, we do not enforce strict pixel-wise matching. Instead, we aim for a weaker consistency that preserves beneficial stylization while filtering meaningless interference.

To reduce the influence of outliers introduced by noise and irrelevant details, we adopt LogMSE for high-frequency components. LogMSE down-weights large errors while remaining sensitive to small-to-moderate discrepancies, which helps refine transferable textures\cite{liu2025distillingcrossmodalknowledgefeature}.

We therefore construct a weighted composite objective by augmenting the base reconstruction loss with a frequency-domain supervision term:

\begin{align}
\mathcal{L}_{\text{stage2}} &= \mathcal{L}_{\text{base}} \nonumber + \lambda_{\text{freq}} \cdot \sum_{k} \Big( w_{\text{low}} \cdot \big\|\hat{\mathbf{z}}^{(k)}_{\text{low}} - \mathbf{z}^{(k)}_{\text{target, low}} \big\|^2_2 \nonumber\\
&\qquad + w_{\text{high}} \cdot \log\big(1+\big\|\hat{\mathbf{z}}^{(k)}_{\text{high}} - \mathbf{z}^{(k)}_{\text{target, high}} \big\|^2_2 \big)\Big).\label{eq:stage2_loss}
\end{align}
Here, $w(t)$ is the timestep weight given by the scheduler during training. $\lambda_{\text{freq}}$ controls the overall strength of frequency supervision. $w_{\text{low}}$ and $w_{\text{high}}$ balance low-frequency structure preservation against high-frequency detail reconstruction. We typically set $w_{\text{high}}>w_{\text{low}}$ to prioritize fine-grained high-frequency refinement.

This staged design, together with multi-scale frequency supervision in Stage 2, improves high-frequency detail reconstruction while preserving global structure.

\subsection{Pixel-space Reward Learning}
\label{sec:reward}
Even after learning a content–style mapping in latent space through semantic–frequency disentanglement, a key challenge remains: latent-space optimization is not perfectly aligned with human perception in pixel space\cite{Gao_2025_ICCV}. Latent reconstruction objectives do not fully capture pixel-level perceptual quality. A solution optimal in latent space may therefore not yield the best decoded perception in terms of content faithfulness, style consistency, or detail sharpness.

To bridge this gap, we incorporate a reward-guided learning mechanism throughout training. A set of multi-dimensional reward models operates on decoded images in pixel space. These models evaluate generated outputs and provide perceptual feedback signals, enabling direct and fine-grained optimization of visual quality and content–style alignment.

Reward-guided learning runs alongside the main training process. Let $S_{\text{total}}$ denote the total training steps and $S_{\text{warmup}}$ the number of Stage 1 steps (default $S_{\text{warmup}} = 0.6\cdot S_{\text{total}}$). For each mini-batch, we determine the current stage based on step index $s$ and compute the corresponding latent-space loss $\mathcal{L}_{\text{latent}}(s)$.

\begin{equation}
\mathcal{L}_{\text{latent}}(s) =
\begin{cases}
\mathcal{L}_{\text{stage1}}, & \text{if } s \leq S_{\text{warmup}},\\
\mathcal{L}_{\text{stage2}}, & \text{otherwise}.
\end{cases}
\label{eq:latent_loss_switch}
\end{equation}

In parallel, the reward module is activated. We define a multi-dimensional reward function in pixel space, $R(\hat{\mathbf{I}},\mathbf{I}_{\text{target}},\mathbf{I}_{\text{content}},\mathbf{I}_{\text{style}})$, to evaluate the decoded image $\hat{\mathbf{I}}$. The reward aggregates several specialized signals. The content-faithfulness reward $R_c$ measures semantic similarity between the generated image and the content reference via CLIP-I\cite{pmlr-v139-radford21a} image encoder, encouraging preservation of key content. The style-alignment reward $R_s$ measures stylistic similarity between the generated image and the style reference via CSD\cite{somepalli2024measuring}, encouraging accurate capture and transfer of artistic style. The total reward is a weighted sum of these core rewards together with auxiliary signals including an LPIPS-based perceptual reward and a discriminator-based adversarial reward:

\begin{equation}
R_{\text{total}} = \omega_c R_c + \omega_s R_s + \sum_i \omega_i R_i.
\label{eq:reward_total}
\end{equation}

Reward learning remains active in both stages: outputs from either Stage 1 or Stage 2 are decoded into pixel space, scored by the reward models, and optimized with a policy-gradient-style objective. To avoid confusion with timestep $t$, we denote the step-wise advantage by $\mathcal{A}(s)$. The pixel-space reward loss is:

\begin{align}
\mathcal{L}_{\text{pixel}}
&= -\mathbb{E}_{(\mathbf{z}^{(t)}_{\text{target}},t,\mathbf{z}^{(t)}_{\text{content}},\mathbf{z}^{(t)}_{\text{style}})\sim\mathcal{D}}
\left[ \mathcal{A}(s)\cdot \log \pi_\theta\!\left(\hat{\mathbf{I}}\,\middle|\,\mathbf{z}^{(t)}_{\text{target}},t,\mathbf{z}^{(t)}_{\text{content}},\mathbf{z}^{(t)}_{\text{style}} \right) \right].
\label{eq:pixel_loss}
\end{align}
Here, $\pi_\theta(\cdot)$ denotes the conditional generative distribution induced by the denoiser (without committing to a specific noise parameterization). The advantage is defined as:

\begin{equation}
\mathcal{A}(s)=R_{\text{total}}\!\left(\hat{\mathbf{I}},\mathbf{I}_{\text{target}},\mathbf{I}_{\text{content}},\mathbf{I}_{\text{style}}\right)-b\!\left(\mathbf{z}^{(t)}_{\text{target}},t,\mathbf{z}^{(t)}_{\text{content}},\mathbf{z}^{(t)}_{\text{style}}\right).
\label{eq:advantage}
\end{equation}
$b(\cdot)$ is the baseline function. Finally, we combine the latent-space loss and the pixel-space reward loss with a weighting coefficient to form the overall training objective:

\begin{equation}
\mathcal{L}_{\text{total}}(s)=\mathcal{L}_{\text{latent}}(s)+\lambda_{\text{pixel}}\cdot \mathcal{L}_{\text{pixel}}(s).
\label{eq:total_objective}
\end{equation}

\section{Experiments}

\subsection{Experimental Settings}
\subsubsection{Setup.}
We adopt Qwen-Image-Edit-2509\cite{wu2025qwenimagetechnicalreport} as the pretrained model. Through the data generation pipeline detailed in the appendix, we construct a training dataset containing 40,000 high-quality four-tuples for model training. Training is conducted on NVIDIA A100 GPUs in two stages: Stage 1 fine-tunes for 12,000 steps on 2 GPUs with a learning rate of $5\times 10^{-6}$ and batch size 1. Stage 2 fine-tunes for 8,000 steps on 2 GPUs with per-GPU batch size 1 (total batch size 2), using the same learning rate.

\subsubsection{Benchmark.}
For fair and comprehensive evaluation, we construct the CSG-Bench benchmark by sampling and augmenting public datasets such as OmniConsistency\cite{Song2025OmniConsistencyLS} and OmniStyle\cite{Wang_2025_CVPR}. The test set contains 1,922 content images spanning portraits, architecture, landscapes, cartoons, objects, food, and animals. For styles, we randomly select 6 out of 20 predefined style types, including 3D Chibi, American Cartoon, clay, Ghibli, Paper cutting, and Vector and curate 100 high-quality, clear reference images for each style. Evaluation results on OmniConsistency-bench are provided in the appendix.

\subsubsection{Evaluation Metrics.}
To comprehensively evaluate generation performance, we adopt a multi-dimensional metric suite covering two core aspects. For content consistency\cite{chen2025consislora}, we report CLIP-I\cite{pmlr-v139-radford21a}, DINO\cite{oquab2024dinov}, and DreamSim\cite{10.5555/3666122.3668330} as complementary measures. For style consistency\cite{song20253sgenunifiedsubjectstyle}, we use FID\cite{10.5555/3295222.3295408}, CLIP-T\cite{pmlr-v139-radford21a}, and CSD\cite{somepalli2024measuring} to assess style alignment from multiple perspectives.

\subsubsection{Baselines.}
We compare UniCSG against eight representative and state-of-the-art baselines: OmniConsistency\cite{Song2025OmniConsistencyLS}, OmniStyle\cite{Wang_2025_CVPR}, flux1.Kontext-dev\cite{labs2025flux1kontextflowmatching}, Nano-banana\cite{comanici2025gemini25pushingfrontier}, DreamOmni2\cite{xia2025dreamomni2multimodalinstructionbasedediting}, OmniGen2\cite{wu2025omnigen2explorationadvancedmultimodal}, BAGEL\cite{deng2025bagel}, and Ovis-U1\cite{wang2025ovisu1}.

\begin{table*}[!t]
\centering
\caption{Quantitative results for text-guided style transfer on CSG-Bench.}
\label{tab:csgbench_text}
\footnotesize
\renewcommand{\arraystretch}{1.4}
\setlength{\tabcolsep}{3.5pt}
\begin{tabular}{lcccccc}
\toprule
\multirow{2}{*}{\textbf{Method}}
& \multicolumn{3}{c}{\textbf{Style consistency}}
& \multicolumn{3}{c}{\textbf{Content consistency}} \\
\cmidrule(lr){2-4} \cmidrule(lr){5-7}
& FID $\downarrow$ & CSD $\uparrow$ & CLIP-T $\uparrow$
& CLIP-I $\uparrow$ & DINO $\uparrow$ & DreamSim $\uparrow$ \\
\midrule
OmniConsistency\cite{Song2025OmniConsistencyLS} & \second{117.079} & 0.503 & 0.266 & 0.705 & 0.487 & 0.712 \\
OmniStyle\cite{Wang_2025_CVPR} & 134.415 & 0.275 & 0.247 & 0.737 & 0.620 & 0.715 \\
flux1.Kontext-dev\cite{labs2025flux1kontextflowmatching} & 120.649 & 0.430 & 0.251 & 0.772 & 0.588 & 0.741 \\
Nano-banana\cite{comanici2025gemini25pushingfrontier} & 128.162 & 0.534 & 0.249 & \second{0.792} & \second{0.644} & \second{0.791} \\
DreamOmni2\cite{xia2025dreamomni2multimodalinstructionbasedediting} & 120.697 & 0.388 & 0.252 & 0.757 & 0.531 & 0.704 \\
OmniGen2\cite{wu2025omnigen2explorationadvancedmultimodal} & 121.355 & 0.448 & 0.260 & 0.733 & 0.547 & 0.733 \\
BAGEL\cite{deng2025bagel} & 132.504 & 0.525 & 0.265 & 0.696 & 0.425 & 0.638 \\
Ovis-U1\cite{wang2025ovisu1} & 136.033 & \best{0.573} & \best{0.275} & 0.702 & 0.400 & 0.653 \\
\midrule
Qwen-Image-Edit\cite{wu2025qwenimagetechnicalreport} (Base) & 120.329 & 0.489 & 0.251 & 0.772 & 0.617 & 0.761 \\
UniCSG (Ours) & \best{113.940} & \second{0.541} & \second{0.267} & \best{0.797} & \best{0.701} & \best{0.816} \\
\bottomrule
\end{tabular}
\end{table*}

\begin{table*}[!t]
\centering
\caption{Quantitative results for reference-guided style transfer on CSG-Bench.}
\label{tab:csgbench_ref}
\footnotesize
\renewcommand{\arraystretch}{1.4}
\setlength{\tabcolsep}{3.5pt}
\begin{tabular}{lcccccc}
\toprule
\multirow{2}{*}{\textbf{Method}}
& \multicolumn{3}{c}{\textbf{Style consistency}}
& \multicolumn{3}{c}{\textbf{Content consistency}} \\
\cmidrule(lr){2-4} \cmidrule(lr){5-7}
& FID $\downarrow$ & CSD $\uparrow$ & CLIP-T $\uparrow$
& CLIP-I $\uparrow$ & DINO $\uparrow$ & DreamSim $\uparrow$ \\
\midrule
OmniConsistency\cite{Song2025OmniConsistencyLS} & 91.140 & 0.635 & 0.269 & 0.666 & 0.408 & 0.679 \\
OmniStyle\cite{Wang_2025_CVPR} & 129.775 & 0.393 & 0.256 & 0.732 & \best{0.615} & 0.670 \\
flux1.Kontext-dev\cite{labs2025flux1kontextflowmatching} & 91.785 & 0.560 & 0.255 & 0.690 & 0.498 & 0.685 \\
Nano-banana\cite{comanici2025gemini25pushingfrontier} & 127.868 & 0.522 & 0.250 & 0.745 & 0.586 & \second{0.748} \\
DreamOmni2\cite{xia2025dreamomni2multimodalinstructionbasedediting} & 112.960 & 0.420 & 0.242 & \best{0.784} & \second{0.606} & \second{0.748} \\
OmniGen2\cite{wu2025omnigen2explorationadvancedmultimodal} & 90.270 & 0.622 & \second{0.271} & 0.624 & 0.409 & 0.624 \\
BAGEL\cite{deng2025bagel} & \best{72.070} & 0.511 & 0.266 & 0.503 & 0.212 & 0.500 \\
Ovis-U1\cite{wang2025ovisu1} & 101.215 & \second{0.703} & \best{0.277} & 0.552 & 0.274 & 0.555 \\
\midrule
Qwen-Image-Edit\cite{wu2025qwenimagetechnicalreport} (Base) & 87.922 & 0.577 & 0.253 & 0.570 & 0.328 & 0.565 \\
UniCSG (Ours) & \second{87.320} & \best{0.731} & \second{0.271} & \second{0.760} & 0.597 & \best{0.762} \\
\bottomrule
\end{tabular}
\end{table*}

\subsection{Experimental Results}
\subsubsection{Quantitative Evaluation.}
In all tables, we highlight the best value in \best{red} and the second-best in \second{blue} in each column. Quantitative comparisons are summarized in Tables~\ref{tab:csgbench_text} and~\ref{tab:csgbench_ref}.

In the text-guided setting, UniCSG achieves a favorable trade-off between style consistency and content consistency. Although style-consistency metrics are slightly lower than Ovis-U1\cite{wang2025ovisu1}, it consistently preserves content structure while maintaining competitive stylization strength. Moreover, the FID score of 113.9397 outperforms multiple baselines (e.g., OmniConsistency\cite{Song2025OmniConsistencyLS}), indicating that the distribution of generated images is closer to the target style distribution.

In the reference-guided setting, UniCSG likewise achieves a strong style-content trade-off. A high CSD score indicates effective style transfer, and a high DreamSim score indicates improved content preservation with reduced reference-content leakage. DINO and CLIP-T scores are comparable to leading models, suggesting competitive visual quality and semantic content retention even when some baselines achieve lower FID at the cost of weaker content control.

\begin{figure*}[!t]
\centering
\includegraphics[width=\textwidth]{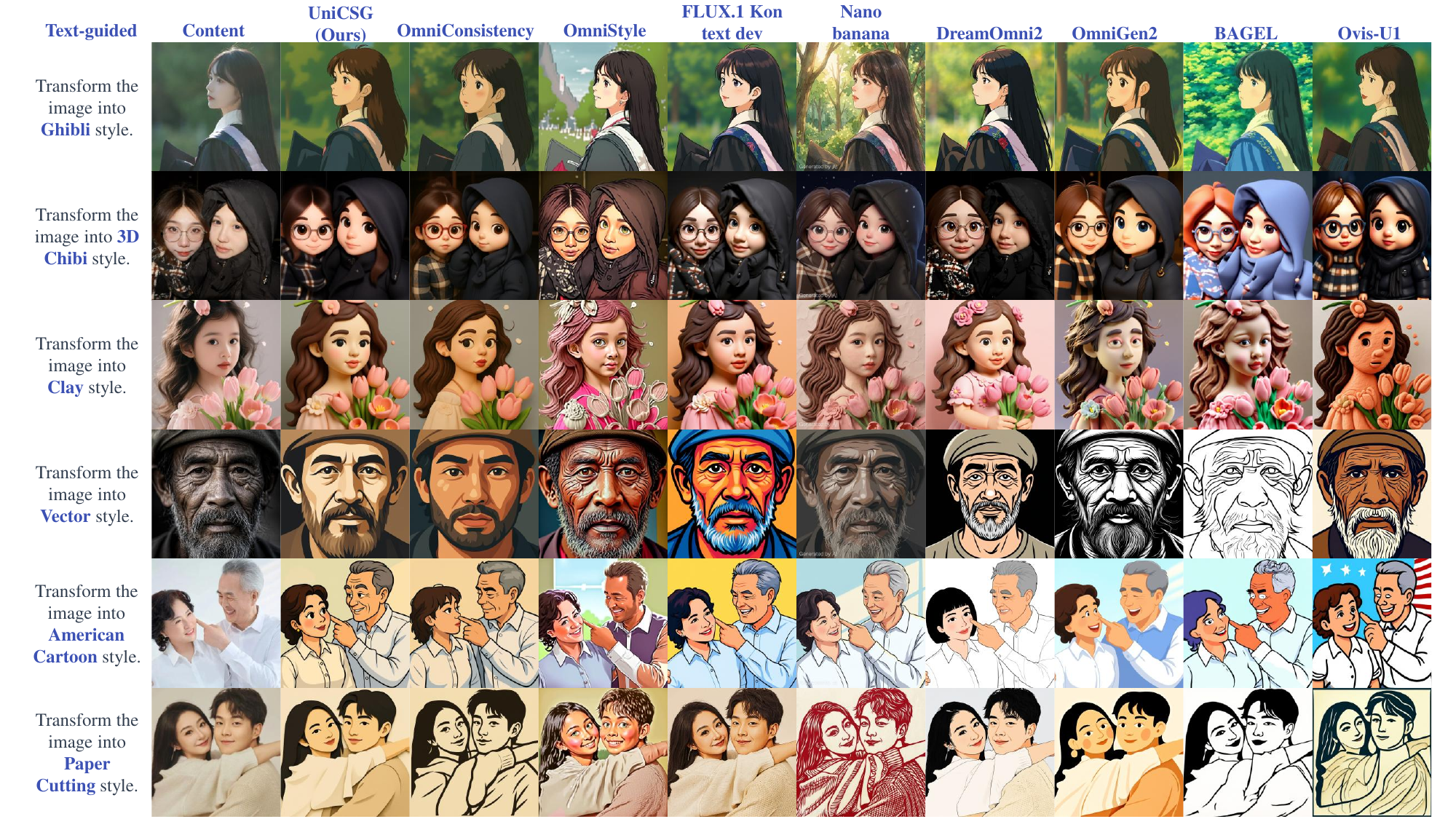}
\caption{Qualitative results for text-guided style transfer on CSG-Bench.}
\label{fig:qual_results_text}
\end{figure*}

\begin{figure*}[!t]
\centering
\includegraphics[width=\textwidth]{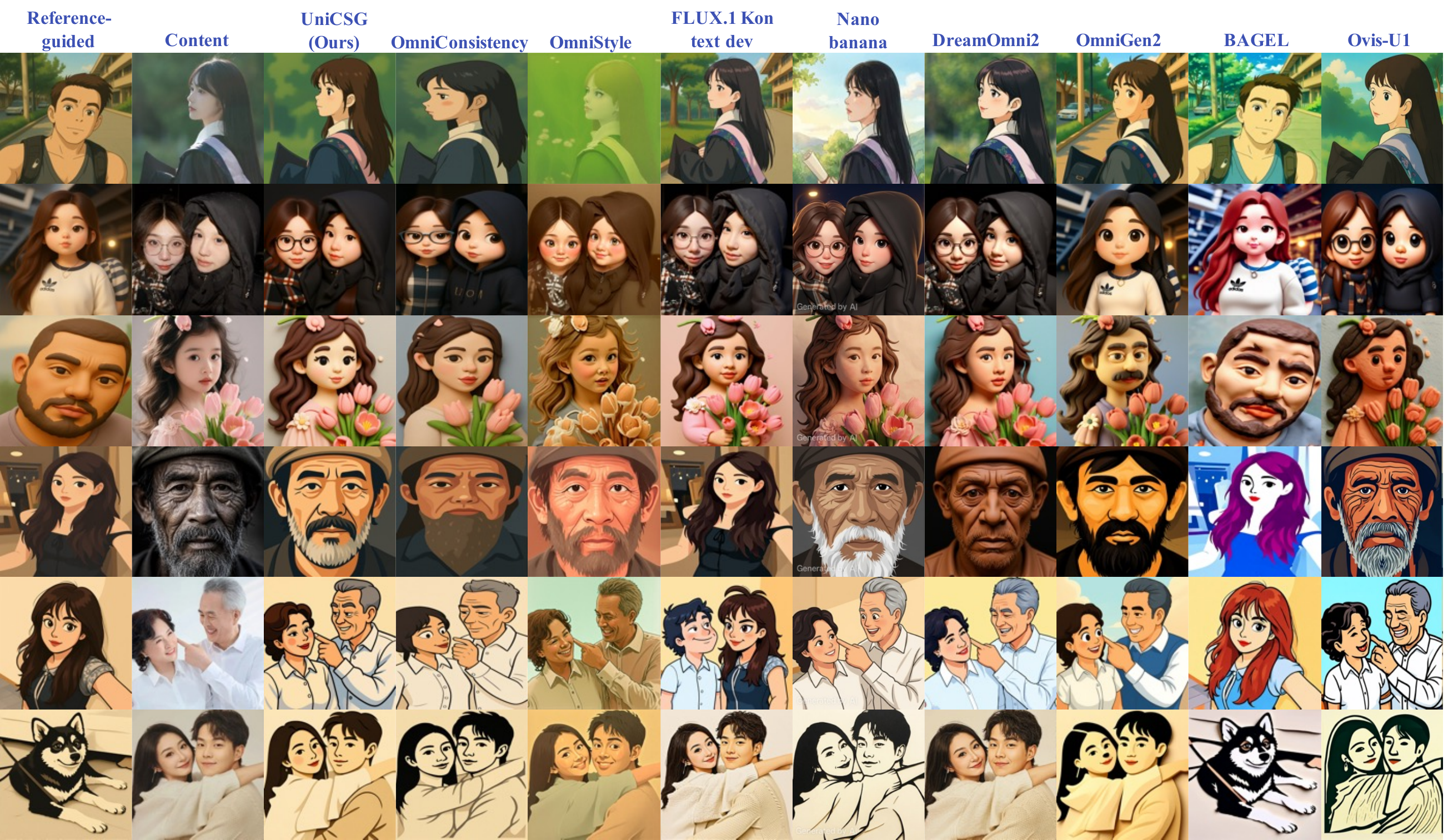}
\caption{Qualitative results for reference-guided style transfer on CSG-Bench.}
\label{fig:qual_results_ref}
\end{figure*}

\subsubsection{Qualitative Evaluation.}
Representative visual comparisons are shown in Fig.~\ref{fig:qual_results_text} and Fig.~\ref{fig:qual_results_ref}.

In the text-guided setting, UniCSG preserves content well and produces strong details. Fine-grained inspection shows better preservation of hairstyle, semantic identity, object count, and background layout consistency, while still maintaining clear stylization. We observe slightly weaker stylization strength than OmniConsistency\cite{Song2025OmniConsistencyLS} and Ovis-U1\cite{wang2025ovisu1} in some cases, which may contribute to the lower style-consistency scores.

In the reference-guided setting, our results differ substantially from the text-guided outputs, indicating that the model can learn style characteristics from the reference image. Meanwhile, BAGEL\cite{deng2025bagel} suffers from severe reference-content leakage, making its style distribution closer to the original content distribution. OmniStyle\cite{Wang_2025_CVPR} and DreamOmni2\cite{xia2025dreamomni2multimodalinstructionbasedediting} exhibit transfer failures in some cases, producing results closer to the original content images. Compared with these baselines, UniCSG better preserves semantic identity and scene structure while reducing the unintended transfer of reference-specific content.

\subsubsection{User Study.}

We conduct a user study to further demonstrate UniCSG's advantages. Thirty evaluators assess 48 image cases covering both text-guided and reference-guided tasks. For each case, evaluators select the best result among nine models in terms of content consistency and style consistency. This human evaluation complements automatic metrics, which may not fully reflect subtle structural drift or reference-content leakage under fine-grained inspection. The preference statistics are reported in Table~\ref{tab:user_study}, where UniCSG receives higher user preference in both settings.

\begin{table*}[!t]
\centering
\caption{User preference rates(\%) for content and style consistency across methods.}
\label{tab:user_study}
\footnotesize
\renewcommand{\arraystretch}{1.4}
\setlength{\tabcolsep}{3.5pt}
\begin{tabular}{lcccc}
\toprule
\multirow{2}{*}{\textbf{Method}} & \multicolumn{2}{c}{\textbf{Text-guided}} & \multicolumn{2}{c}{\textbf{Reference-guided}} \\
\cmidrule(lr){2-3} \cmidrule(lr){4-5}
& Content $\uparrow$ & Style $\uparrow$ & Content $\uparrow$ & Style $\uparrow$ \\
\midrule
OmniConsistency\cite{Song2025OmniConsistencyLS} & 6.9 & 9.9 & 3.2 & 5.6 \\
OmniStyle\cite{Wang_2025_CVPR} & 1.8 & 1.2 & 2.0 & 2.0 \\
flux1.Kontext-dev\cite{labs2025flux1kontextflowmatching} & \second{26.6} & 22.0 & 11.5 & 7.1 \\
Nano-banana\cite{comanici2025gemini25pushingfrontier} & 25.8 & \second{22.8} & \second{30.4} & \second{24.2} \\
DreamOmni2\cite{xia2025dreamomni2multimodalinstructionbasedediting} & 4.0 & 3.6 & 11.3 & 8.7 \\
OmniGen2\cite{wu2025omnigen2explorationadvancedmultimodal} & 3.4 & 5.8 & 0.2 & 3.2 \\
BAGEL\cite{deng2025bagel} & 0.6 & 1.8 & 0.0 & 0.8 \\
Ovis-U1\cite{wang2025ovisu1} & 1.4 & 2.8 & 3.2 & 3.8 \\
\midrule
UniCSG (Ours) & \best{29.6} & \best{30.2} & \best{38.3} & \best{44.6} \\
\bottomrule
\end{tabular}
\end{table*}

\subsubsection{Ablation Study.}

\begin{figure*}[!t]
\centering
\includegraphics[width=\textwidth]{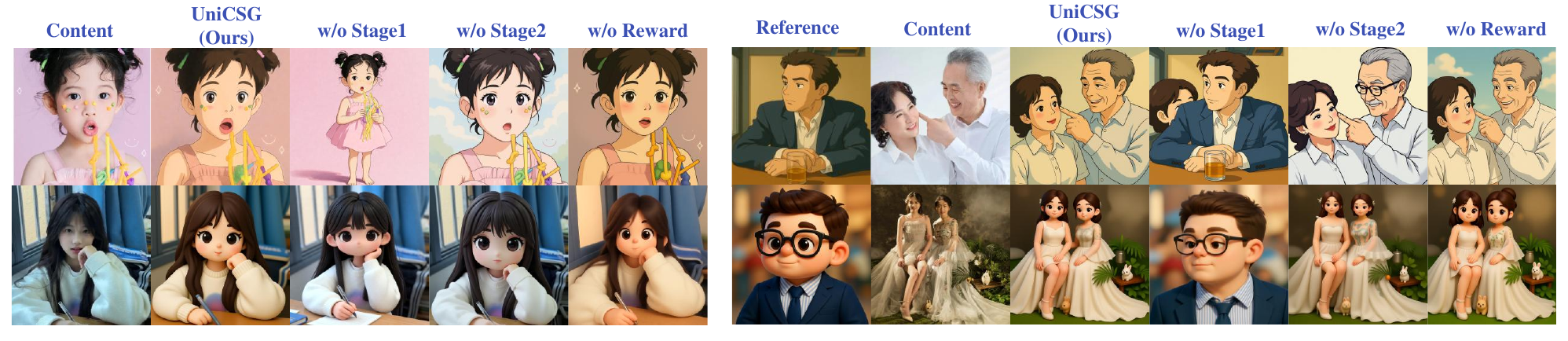}
\caption{Ablation study results. The first row shows the Ghibli style and the second row shows the 3D Chibi style. Left: text-guided setting. Right: reference-guided setting.}
\label{fig:ablation_results}
\end{figure*}

\begin{table*}[!t]
\centering
\caption{Ablation results for text-guided style transfer on CSG-Bench.}
\label{tab:ablation_text}
\footnotesize
\renewcommand{\arraystretch}{1.4}
\setlength{\tabcolsep}{3.5pt}
\begin{tabular}{lcccccc}
\toprule
\multirow{2}{*}{\textbf{Method}} & \multicolumn{3}{c}{\textbf{Style consistency}} & \multicolumn{3}{c}{\textbf{Content consistency}} \\
\cmidrule(lr){2-4} \cmidrule(lr){5-7}
& FID $\downarrow$ & CSD $\uparrow$ & CLIP-T $\uparrow$ & CLIP-I $\uparrow$ & DINO $\uparrow$ & DreamSim $\uparrow$ \\
\midrule
UniCSG (Ours) & \best{113.940} & \best{0.541} & \best{0.267} & \best{0.797} & \best{0.701} & \best{0.816}\\
\midrule
w/o Stage 1 & \second{121.855} & \second{0.524} & \second{0.258} & 0.748 & 0.590 & 0.763 \\
w/o Stage 2 & 124.718 & 0.516 & 0.255 & 0.755 & \second{0.630} & 0.770 \\
w/o Reward  & 123.331 & 0.521 & 0.255 & \second{0.758} & 0.622 & \second{0.775} \\
\bottomrule
\end{tabular}
\end{table*}

\begin{table*}[!t]
\centering
\caption{Ablation results for reference-guided style transfer on CSG-Bench.}
\label{tab:ablation_ref}
\footnotesize
\renewcommand{\arraystretch}{1.4}
\setlength{\tabcolsep}{3.5pt}
\begin{tabular}{lcccccc}
\toprule
\multirow{2}{*}{\textbf{Method}} & \multicolumn{3}{c}{\textbf{Style consistency}} & \multicolumn{3}{c}{\textbf{Content consistency}} \\
\cmidrule(lr){2-4} \cmidrule(lr){5-7}
& FID $\downarrow$ & CSD $\uparrow$ & CLIP-T $\uparrow$ & CLIP-I $\uparrow$ & DINO $\uparrow$ & DreamSim $\uparrow$ \\
\midrule
UniCSG (Ours) & \best{87.320} & \best{0.731} & \best{0.271} & \best{0.760} & \best{0.597} & \best{0.762} \\
\midrule
w/o Stage 1 & \second{90.708} & 0.645 & \second{0.266} & 0.716 & 0.417 & 0.691 \\
w/o Stage 2 & 92.317 & 0.619 & 0.262 & 0.714 & \second{0.452} & 0.701 \\
w/o Reward  & 92.335 & \second{0.661} & 0.264 & \second{0.723} & 0.451 & \second{0.706} \\
\bottomrule
\end{tabular}
\end{table*}
We visualize ablation trends in Fig.~\ref{fig:ablation_results} and report detailed numbers in Tables~\ref{tab:ablation_text} and~\ref{tab:ablation_ref}.
We conduct ablations on three components: Stage 1, Stage 2, and the reward model. Removing Stage 1 (low-frequency training with conditioning corruption) significantly degrades content-consistency metrics: in the text-guided setting, content becomes uncontrolled with hallucinations, while in the reference-guided setting, severe reference-content leakage occurs, demonstrating the importance of learning low-frequency structure for content constraints. Removing Stage 2 (frequency supervision) reduces style-consistency performance, including weaker geometric deformation and less accurate color tones, suggesting that high-frequency refinement improves stylization by recovering fine-grained textures and geometric cues. Removing the reward model yields a smaller but consistent drop, consistent with its role in perceptual alignment, particularly refining background and clothing details. Gains are larger in the reference-guided setting, indicating that UniCSG is particularly effective for reference-based stylization.

\section{Discussion and Limitations}
We discuss the generality and limitations of UniCSG from three perspectives: generalization, style transfer granularity, and efficiency.

\begin{figure*}[!t]
\centering
\includegraphics[width=0.95\textwidth]{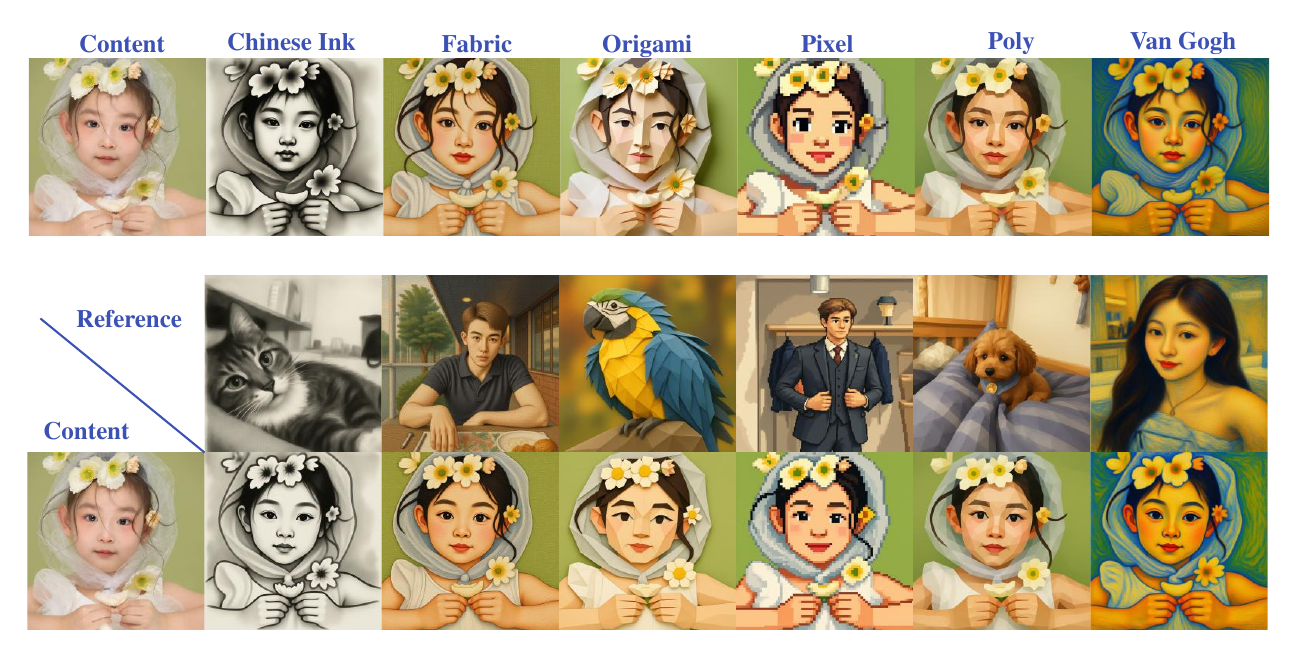}
\caption{Qualitative results on unseen styles during training.}
\label{fig:csgbench_seen_unseen}
\end{figure*}

\begin{table*}[!t]
\centering
\caption{Quantitative results under seen and unseen settings on CSG-Bench.}
\label{tab:csgbench_seen_unseen}
\footnotesize
\renewcommand{\arraystretch}{1.4}
\setlength{\tabcolsep}{3.5pt}
\begin{tabular}{lcccccc}
\toprule
\multirow{2}{*}{\textbf{Task-Style setting}} & \multicolumn{3}{c}{\textbf{Style consistency}} & \multicolumn{3}{c}{\textbf{Content consistency}} \\
\cmidrule(lr){2-4} \cmidrule(lr){5-7}
& FID $\downarrow$ & CSD $\uparrow$ & CLIP-T $\uparrow$ & CLIP-I $\uparrow$ & DINO $\uparrow$ & DreamSim $\uparrow$ \\
\midrule
Text-Seen & 113.940 & 0.541 & 0.267 & 0.797 & 0.701 & 0.816\\
Text-Unseen & 118.303 & 0.541 & 0.259 & 0.734 & 0.546 & 0.726 \\
\midrule
Reference-Seen & 87.320 & 0.731 & 0.271 & 0.760 & 0.597 & 0.762 \\
Reference-Unseen  & 113.532 & 0.701 & 0.257 & 0.728 & 0.536 & 0.718 \\
\bottomrule
\end{tabular}
\end{table*}

\subsubsection{Generalization to unseen styles.}

Our staged training explicitly separates semantic consistency (Stage~1) from texture stylization (Stage~2), and randomly sampling from the style pool during training exposes the model to diverse styles. Together, these design choices improve generalization to unseen style. Fig.~\ref{fig:csgbench_seen_unseen} and Table~\ref{tab:csgbench_seen_unseen} show that performance degrades on unseen styles, which is expected because the unseen setting provides a harder transfer scenario without exposure to that style category during training. Nevertheless, UniCSG retains relatively strong content-related metrics, suggesting that it learns transferable content-preservation capability rather than merely overfitting to seen-style distributions.

\subsubsection{Image-level vs. category-level stylization.}

\begin{figure*}[!t]
\centering
\includegraphics[width=\textwidth]{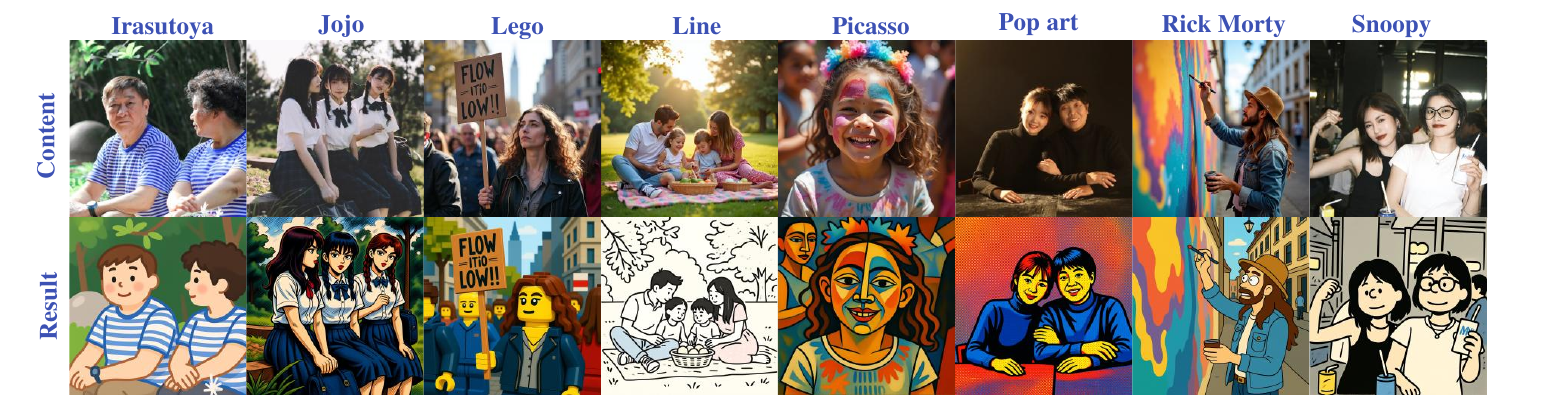}
\caption{Qualitative results across diverse style categories.}
\label{fig:other_styles}
\end{figure*}

Our style pool covers diverse artistic categories, including atmosphere and palette changes, stroke- and texture-driven rendering, and geometric or shape-deforming stylization. This diversity highlights a key distinction from image-level transfer, which is often closer to atmosphere or appearance translation tied to a single reference image. In contrast, UniCSG targets category-level stylization: beyond color tone and rendering patterns, it also learns transferable stylistic transformations shared across a style category, including geometric deformation and structural restylization, while preserving the content semantics of the source image. As a result, the method is better suited to content-constrained stylization settings where avoiding reference-specific leakage is as important as achieving strong visual style transfer. Fig.~\ref{fig:other_styles} illustrates representative results across different style categories.

\subsubsection{Efficiency.}

UniCSG is built on the Qwen-Image-Edit-2509 backbone and further accelerated with Qwen-Image-Lightning, reducing per-image inference latency to approximately 4s for text-guided tasks and 10s for reference-guided tasks at 1024 resolution.

\section{Conclusion}
We propose UniCSG, a unified training framework for high-fidelity, content-constrained, style-driven generation. With a staged design semantic disentanglement, frequency-aware reconstruction, and pixel-space reward learning. UniCSG improves style-transfer stability and fine-detail quality while preserving content structure and substantially mitigating reference-content leakage. Future work will explore stronger cross-domain generalization and more efficient reward modeling to further improve controllable generation in complex scenes.

\bibliographystyle{unsrtnat}
\bibliography{custom}

\clearpage
\onecolumn
\beginappendix

\renewcommand{\thetable}{\thesection.\arabic{table}}
\renewcommand{\thefigure}{\thesection.\arabic{figure}}

\setcounter{table}{0}
\setcounter{figure}{0}

\appendix
\section{Details of CSG-Dataset}
\subsection{Dataset Generation Pipeline}

To meet the unique data requirements of high-fidelity style transfer, we design a systematic data generation pipeline. The pipeline constructs high-quality four-tuple training data consisting of a content image (source\_img), a reference style image (ref\_img), a target stylized image (gt\_img), and a description text (text).

\paragraph{Pre-processing stage.}
We preprocess real-world content images and artistic images across diverse styles, including data cleaning and standardized augmentation. We also use similarity-based grouping to distinguish portrait versus scenery themes, ensuring semantic consistency.

\paragraph{Generation stage.}
We generate data through stylization and de-stylization models. The stylization model converts content images into stylized images (gt\_img), while the de-stylization model converts stylized images back into RGB images (source\_img). The reference images (ref\_img) are stylized images that are outputs of the stylization model and inputs to the de-stylization model, and can mutually serve as references. For text construction, an image understanding model processes the input image to generate a content description text, which is then prefixed with task instructions and suffixed with prompt constraints. By combining these elements, we form four-tuples $\langle text, ref\_img, source\_img, gt\_img \rangle$. This design ensures style consistency across the dataset while enabling rolling diversity through cross-referencing between stylized and de-stylized images, significantly enriching the dataset with diverse style-content combinations.

\begin{figure*}[!t]
\centering
\includegraphics[width=0.98\textwidth]{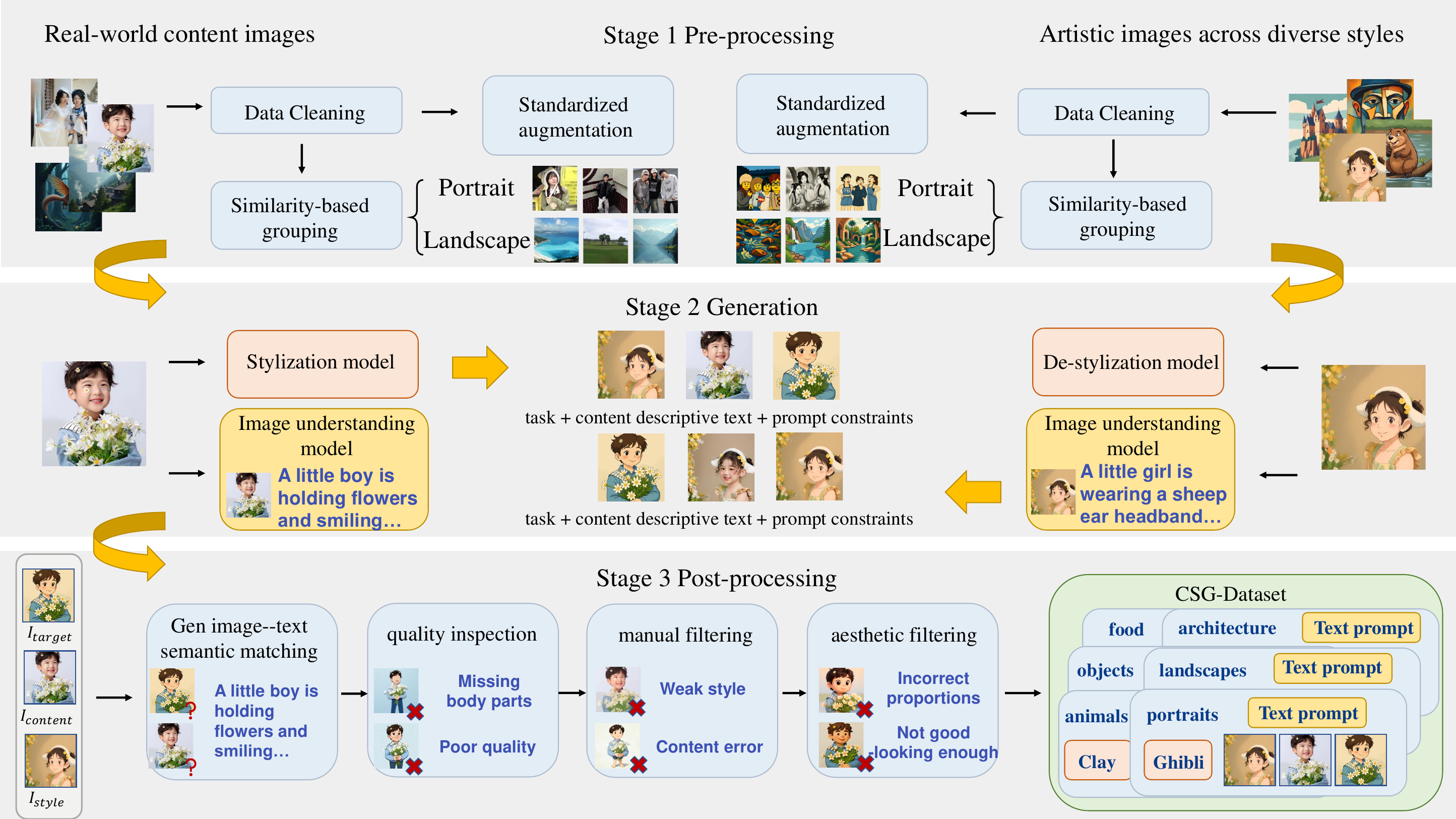}
\caption{Overview of the CSG-Dataset generation pipeline. The pipeline consists of three main stages: (1) Pre-processing: data cleaning and semantic grouping of content and artistic images. (2) Generation: mutual construction between stylization and de-stylization models creates diverse style-content combinations through iterative cycles, with text construction via image understanding. (3) Post-processing: four complementary filtering processes (image--text semantic matching, quality inspection, manual filtering, and aesthetic assessment) ensure high-quality data curation.}
\label{fig:data_pipeline}
\end{figure*}

\paragraph{Post-processing stage.}
The post-processing stage employs four complementary filtering processes to ensure high-quality data curation: image--text semantic matching, quality inspection, manual filtering, and aesthetic assessment. A CLIP-based image--text semantic matching model verifies alignment between generated images and their text descriptions by computing similarity scores, ensuring that the generated results match the text descriptions. We apply a ViT-based quality inspection model to automatically filter results with structural defects such as missing body parts or facial features, as well as quality issues including blur, ghosting, and artifacts. Human annotators perform manual filtering, with each image pair evaluated by at least two annotators. They filter out cases with weak stylization or insignificant de-stylization effects, poor content control such as hallucinated content additions, and only pairs with unanimous approval are retained. An aesthetic assessment model evaluates visual appeal and artistic quality, filtering out images with inappropriate proportions (e.g., oversized heads) or insufficient visual appeal. These four filtering processes work together to significantly improve the overall quality and consistency of the resulting CSG-Dataset. An overview of the complete data generation pipeline is illustrated in Fig.~\ref{fig:data_pipeline}, showing the mutual construction process and filtering stages.

\begin{figure*}
\centering
\includegraphics[width=0.9\textwidth]{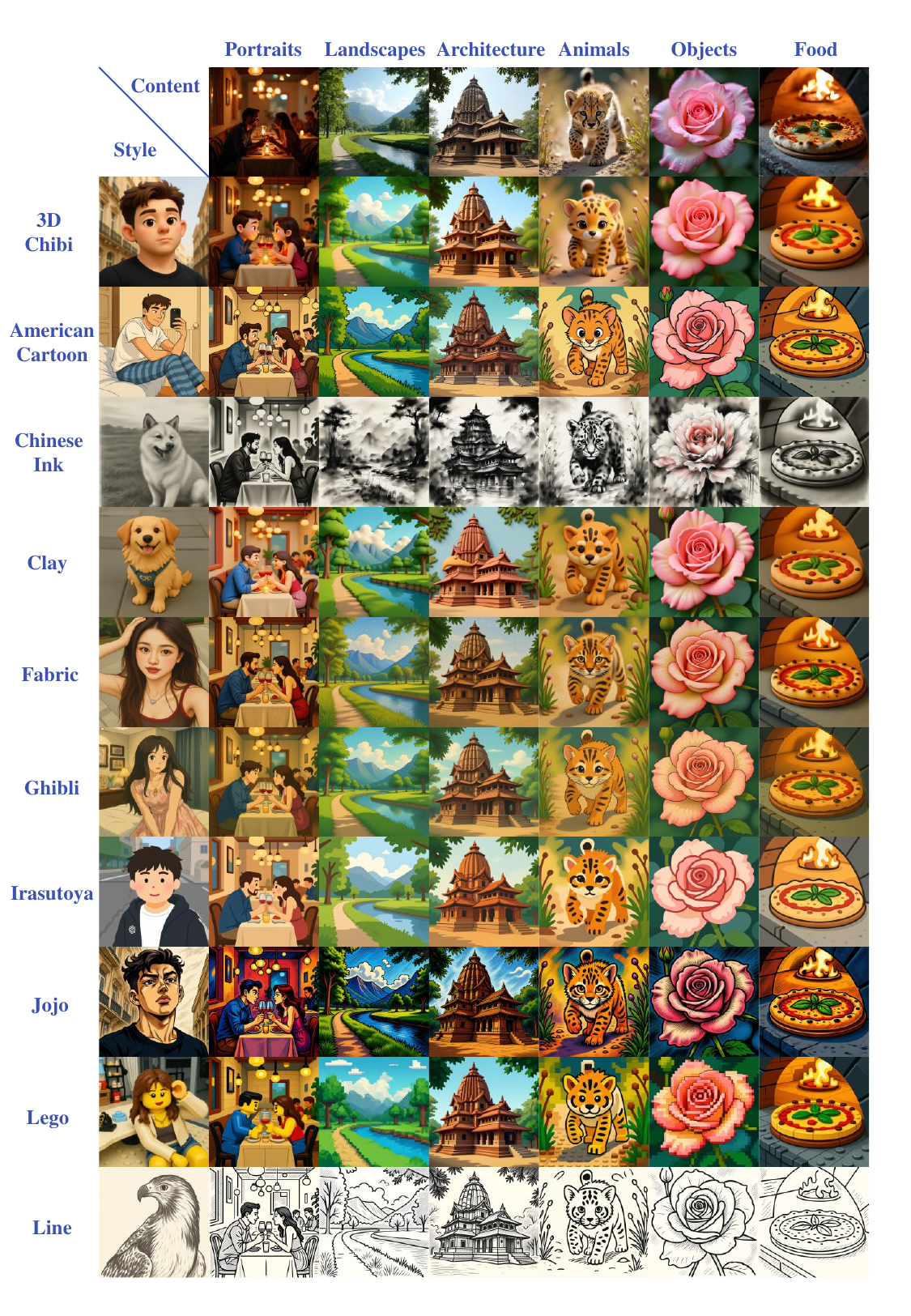}
\caption{Diversity of CSG-Dataset. The figure shows a matrix visualization where rows represent 20 style types (reference images) and columns represent 6 content types (portraits, landscapes, architecture, animals, objects, food). Each cell displays the generated result from our model, demonstrating the comprehensive coverage of style-content combinations in the dataset.}
\label{fig:dataset_diversity}
\end{figure*}

\begin{figure*}
\centering
\includegraphics[width=0.9\textwidth]{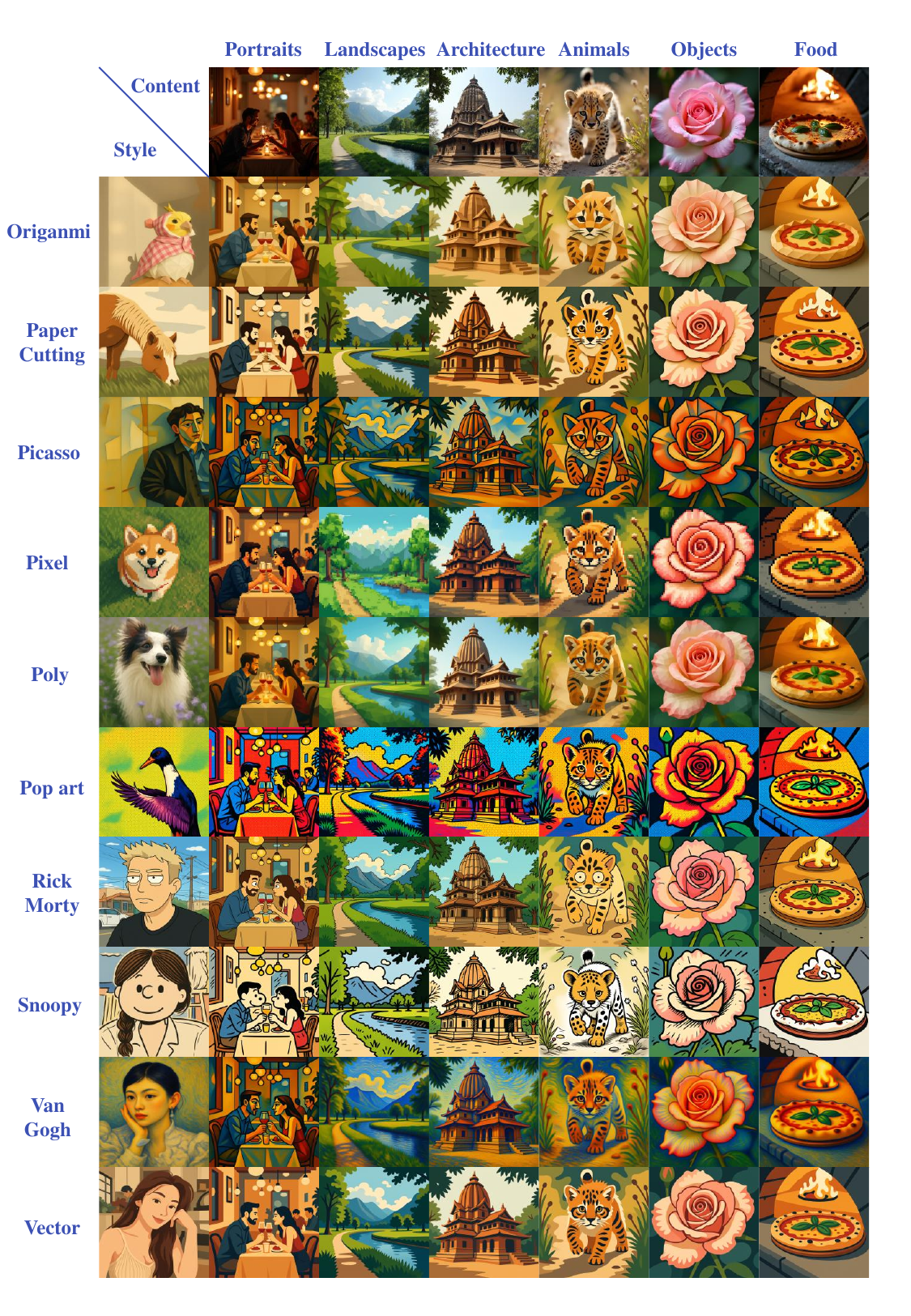}
\caption{Diversity of CSG-Dataset (continued).}
\label{fig:dataset_diversity2}
\end{figure*}

\subsection{Dataset Statistics and Curation}
Through our systematic data generation pipeline, we construct a comprehensive training dataset containing 40,000 high-quality four-tuples (text, ref\_img, source\_img, gt\_img). The dataset covers 20 distinct artistic styles used for training, spanning diverse categories including atmosphere styles (Ghibli, Pop Art, Van Gogh), stroke and texture styles (Paper cutting, Vector, Chinese Ink), and geometric deformation styles (3D Chibi, clay, American Cartoon).

For evaluation purposes, we define 20 styles in total: the 14 training styles (including seen styles evaluation settings), plus 6 additional unseen styles reserved for inference testing. This design enables comprehensive assessment of both in-distribution and out-of-distribution generalization capabilities. The diversity of our dataset is illustrated in Fig.~\ref{fig:dataset_diversity} and Fig.~\ref{fig:dataset_diversity2}, showcasing various content types (portraits, landscapes, architecture, animals, objects, food) paired with different style categories.

\section{Details of CSG-Bench}

\subsection{Metrics}
We adopt a multi-dimensional metric suite. For content consistency, we use CLIP-I, DINO, and DreamSim: CLIP-I measures high-level semantic alignment in CLIP image space; DINO captures fine-grained structural and texture similarity via self-supervised features; DreamSim evaluates perceptual structural similarity aligned with human judgments. For style consistency, we use FID, CLIP-T, and CSD: FID measures distributional distance between generated and target-style images; CLIP-T assesses semantic adherence to style text instructions; CSD directly quantifies artistic-feature similarity to the style reference.

\subsection{Quantitative Evaluation on OmniConsistency-Bench}
OmniConsistency-Bench serves as a subset of our comprehensive CSG-Bench, providing a widely-adopted benchmark for style transfer evaluation. While our main paper has already demonstrated UniCSG's advantages on CSG-Bench, the results presented in Tables~\ref{tab:omnicbench_text} and~\ref{tab:omnicbench_ref} show that these advantages are even more pronounced on this established benchmark subset, further validating the robustness and effectiveness of our method.

\begin{table*}[t]
\centering
\caption{Quantitative results on OmniConsistency-Bench for text-guided style transfer.}
\label{tab:omnicbench_text}
\footnotesize
\renewcommand{\arraystretch}{1.4}
\setlength{\tabcolsep}{3.5pt}
\begin{tabular}{lcccccc}
\toprule
\multirow{2}{*}{\textbf{Method}} & \multicolumn{3}{c}{\textbf{Style consistency}} & \multicolumn{3}{c}{\textbf{Content consistency}} \\
\cmidrule(lr){2-4} \cmidrule(lr){5-7}
& FID $\downarrow$ & CSD $\uparrow$ & CLIP-T $\uparrow$ & CLIP-I $\uparrow$ & DINO $\uparrow$ & DreamSim $\uparrow$ \\

\midrule
OmniConsistency\cite{Song2025OmniConsistencyLS} & 125.679 & 0.527 & \second{0.269} & 0.686 & 0.498 & 0.718 \\
OmniStyle\cite{Wang_2025_CVPR} & 160.028 & 0.284 & 0.248 & 0.697 & 0.591 & 0.729 \\
flux1.Kontext-dev\cite{labs2025flux1kontextflowmatching} & 128.763 & 0.455 & 0.254 & 0.747 & 0.596 & 0.759 \\
Nano-banana\cite{comanici2025gemini25pushingfrontier} & 132.148 & 0.539 & 0.254 & \second{0.756} & \second{0.639} & \second{0.782} \\
DreamOmni2\cite{xia2025dreamomni2multimodalinstructionbasedediting} & \second{125.251} & 0.396 & 0.255 & 0.722 & 0.534 & 0.705 \\
OmniGen2\cite{wu2025omnigen2explorationadvancedmultimodal} & 129.961 & 0.473 & 0.267 & 0.690 & 0.537 & 0.732 \\
BAGEL\cite{deng2025bagel} & 135.621 & 0.552 & 0.268 & 0.672 & 0.440 & 0.641 \\
Ovis-U1\cite{wang2025ovisu1} & 138.369 & \second{0.589} & \best{0.279} & 0.670 & 0.402 & 0.655 \\
\midrule
Qwen-Image-Edit \cite{wu2025qwenimagetechnicalreport}(Base) & 127.815 & 0.490 & 0.252 & 0.734 & 0.622 & 0.754 \\
UniCSG (Ours) & \best{125.067} & \best{0.602} & \second{0.269} & \best{0.770} & \best{0.687} & \best{0.807} \\
\bottomrule
\end{tabular}
\end{table*}

\begin{table*}[t]
\centering
\caption{Quantitative results on OmniConsistency-Bench for reference-guided style transfer.}
\label{tab:omnicbench_ref}
\footnotesize
\renewcommand{\arraystretch}{1.4}
\setlength{\tabcolsep}{3.5pt}
\begin{tabular}{lcccccc}
\toprule
\multirow{2}{*}{\textbf{Method}} & \multicolumn{3}{c}{\textbf{Style consistency}} & \multicolumn{3}{c}{\textbf{Content consistency}} \\
\cmidrule(lr){2-4} \cmidrule(lr){5-7}
& FID $\downarrow$ & CSD $\uparrow$ & CLIP-T $\uparrow$ & CLIP-I $\uparrow$ & DINO $\uparrow$ & DreamSim $\uparrow$ \\

\midrule
OmniConsistency\cite{Song2025OmniConsistencyLS} & 88.282 & 0.669 & 0.272 & 0.623 & 0.401 & 0.679 \\
OmniStyle\cite{Wang_2025_CVPR} & 130.365 & 0.422 & 0.258 & 0.691 & 0.591 & 0.679 \\
flux1.Kontext-dev\cite{labs2025flux1kontextflowmatching} & 92.448 & 0.604 & 0.260 & 0.651 & 0.501 & 0.681 \\
Nano-banana\cite{comanici2025gemini25pushingfrontier} & 121.700 & 0.533 & 0.254 & 0.714 & 0.587 & 0.751 \\
DreamOmni2\cite{xia2025dreamomni2multimodalinstructionbasedediting} & 118.203 & 0.415 & 0.244 & \best{0.759} & \second{0.615} & \second{0.752} \\
OmniGen2\cite{wu2025omnigen2explorationadvancedmultimodal} & 95.784 & 0.630 & 0.273 & 0.625 & 0.455 & 0.661 \\
BAGEL\cite{deng2025bagel} & \best{71.159} & 0.509 & 0.267 & 0.502 & 0.249 & 0.520 \\
Ovis-U1\cite{wang2025ovisu1} & 105.036 & \second{0.710} & \second{0.278} & 0.538 & 0.301 & 0.574 \\
\midrule
Qwen-Image-Edit \cite{wu2025qwenimagetechnicalreport}(Base) & \second{82.796} & 0.540 & 0.258 & 0.547 & 0.330 & 0.569 \\
UniCSG (Ours) & 88.428 & \best{0.759} & \best{0.283} & \second{0.721} & \best{0.616} & \best{0.755} \\
\bottomrule
\end{tabular}
\end{table*}

In the text-guided setting, UniCSG achieves strong performance across both style and content consistency metrics, with particularly notable improvements in content preservation. The results align with the observations on CSG-Bench, where UniCSG excels in maintaining content structure while effectively transferring styles.

In the reference-guided setting, UniCSG demonstrates balanced performance between style alignment and content preservation. While UniCSG's FID score does not rank in the top two, this is because the top two methods (BAGEL and Qwen-Image-Edit) suffer from severe reference-content leakage, as evidenced by their poor content consistency metrics. The results confirm the effectiveness of our staged semantic--frequency disentanglement approach in preventing reference-content leakage, which is a key challenge in reference-guided style transfer. The consistent improvements across both settings validate that our method successfully addresses the content--style entanglement challenge while maintaining high-fidelity generation quality.

\section{Implementation Details}
\subsection{Hyperparameter Settings}

Table~\ref{tab:hyperparams} summarizes the key hyperparameters used in our experiments. The low-pass filter threshold $\tau=0.2$ is selected to retain global structure while filtering high-frequency details. Noise amplification factors $\gamma_{\text{content}}=1.5$ and $\gamma_{\text{style}}=2.0$ create the desired information hierarchy. The probability $p=0.1$ for conditioning replacement balances generalization and training stability. Loss weights are tuned to balance different objectives: semantic losses use $\lambda^{\text{cont}}_{\text{target}}=0.5$, $\lambda^{\text{cont}}_{\text{content}}=0.3$, and $\lambda^{\text{diss}}_{\text{style}}=0.2$. Frequency supervision uses $\lambda_{\text{freq}}=0.1$ with $w_{\text{low}}=1.0$ and $w_{\text{high}}=2.0$ to emphasize high-frequency refinement. The pixel-space reward weight $\lambda_{\text{pixel}}=0.05$ provides gentle guidance without overwhelming the latent-space objectives.

\begin{table}[t]
\centering
\caption{Key hyperparameters for UniCSG training.}
\label{tab:hyperparams}
\footnotesize
\renewcommand{\arraystretch}{1.4}
\setlength{\tabcolsep}{3.5pt}
\begin{tabular}{l@{\hspace{1.5cm}}l}
\toprule
\textbf{Hyperparameter} & \textbf{Value} \\
\midrule
Low-pass filter threshold $\tau$ & 0.2 \\
Noise amplification $\gamma_{\text{content}}$ & 1.5 \\
Noise amplification $\gamma_{\text{style}}$ & 2.0 \\
Conditioning replacement probability $p$ & 0.1 \\
Stage 1/Stage 2 ratio $S_{\text{warmup}}/S_{\text{total}}$ & 0.6 \\
\midrule
\textit{Semantic loss weights:} & \\
\quad $\lambda^{\text{cont}}_{\text{target}}$ & 0.5 \\
\quad $\lambda^{\text{cont}}_{\text{content}}$ & 0.3 \\
\quad $\lambda^{\text{diss}}_{\text{style}}$ & 0.2 \\
\midrule
\textit{Frequency supervision:} & \\
\quad $\lambda_{\text{freq}}$ & 0.1 \\
\quad $w_{\text{low}}$ & 1.0 \\
\quad $w_{\text{high}}$ & 2.0 \\
\midrule
Pixel-space reward weight $\lambda_{\text{pixel}}$ & 0.05 \\
\midrule
Learning rate & $5 \times 10^{-6}$ \\
Batch size (Stage 1) & 1 per GPU \\
Batch size (Stage 2) & 1 per GPU (total 2) \\
Optimizer & AdamW \\
Weight decay & $1 \times 10^{-4}$ \\
\bottomrule
\end{tabular}
\end{table}

\subsection{Network Architecture Details}

We build UniCSG on Qwen-Image-Edit-2509, which uses a DiT-based architecture. The model processes four-tuple inputs $\langle text, ref\_img, source\_img, gt\_img \rangle$ through a unified conditioning mechanism. Content and style images are encoded via the VAE encoder into latent representations. Text prompts are processed through a Qwen2.5-VL. The DiT backbone integrates these conditions through cross-attention layers. For reward learning, we use CLIP-I (ViT-L/14) for content-faithfulness rewards, CSD for style-alignment rewards, LPIPS for perceptual rewards, and a lightweight discriminator for adversarial rewards.

\subsection{Frequency Decomposition Implementation}

We implement multi-scale frequency decomposition using a DoG (Difference of Gaussian) pyramid. Specifically, we apply Gaussian blur with standard deviations $\sigma_k = 2^k$ for $k \in \{0, 1, 2, 3\}$ to create a 4-level pyramid. The low-frequency component at level $k$ is obtained by subtracting the blurred version at level $k+1$ from level $k$. High-frequency components are computed as residuals. This decomposition is performed directly on latent representations to maintain consistency with the training objective in latent space.

\subsection{Conditioning Corruption Schedule}

The corruption strength follows a progressive decay schedule during training. For Stage 1, we linearly decay the noise amplification factors from their initial values ($\gamma_{\text{content}}=1.5$, $\gamma_{\text{style}}=2.0$) to 1.0 over the first 60\% of training steps. This smooth transition prepares the model for Stage 2, where full-detail reconstruction occurs without corruption. The decay schedule is: $\gamma_i(s) = \gamma_i^{\text{init}} \cdot (1 - \frac{s}{S_{\text{warmup}}})$ for $s \leq S_{\text{warmup}}$, and $\gamma_i(s) = 1.0$ for $s > S_{\text{warmup}}$.

\section{More Results}
\subsection{Ablation Study}

\begin{figure*}[!t]
\centering
\includegraphics[width=\textwidth]{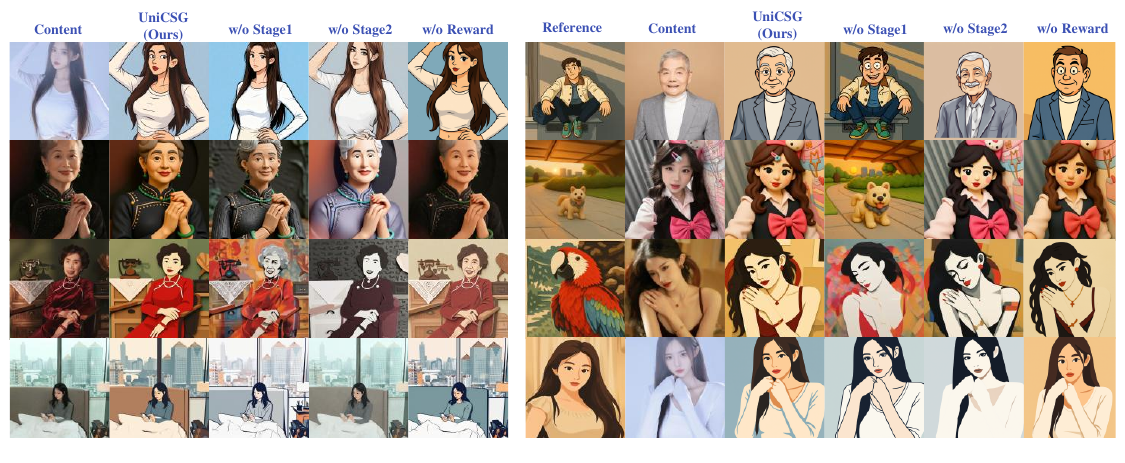}
\caption{Visual ablation study results. The figure shows four rows corresponding to four style types: American Cartoon, clay, Paper cutting, and Vector.}
\label{fig:ablation_visual}
\end{figure*}

We provide visual ablation results that align with the quantitative analysis in the main paper. Fig.~\ref{fig:ablation_visual} shows progressive improvements from individual components across four style types: American Cartoon, clay, Paper cutting, and Vector. Each row demonstrates the contribution of each component to content preservation, style alignment, and overall visual quality.

\subsection{Failure Case Analysis}

\begin{figure}[!t]
\centering
\includegraphics[width=0.9\textwidth]{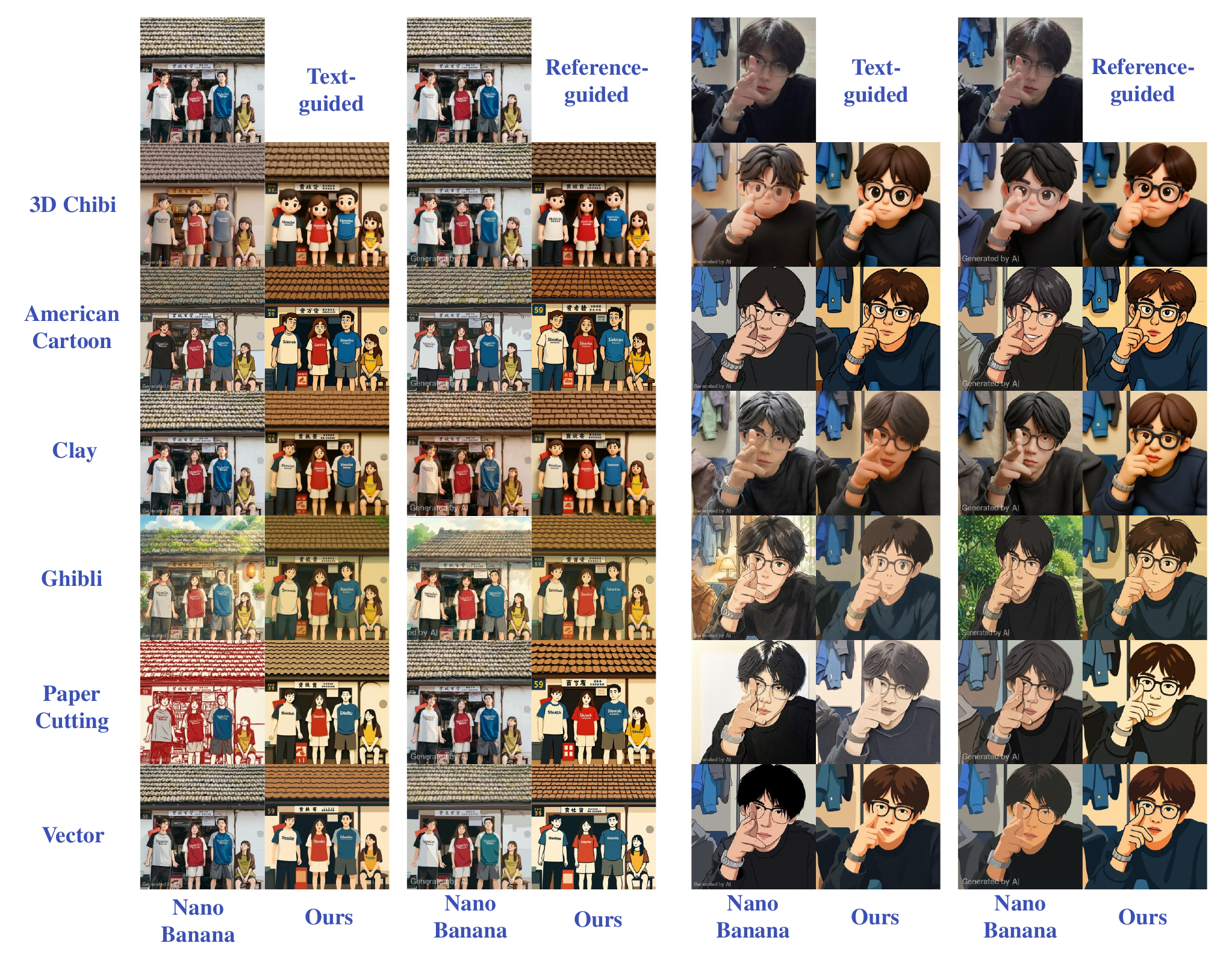}
\caption{Representative failure cases comparing UniCSG and Nano-banana across six styles in both text-guided and reference-guided settings. (a) Text distortion case: comparison of text preservation and overall stylization quality. (b) Hand gesture deformation case: comparison of hand structure preservation across different styles.}
\label{fig:failure_cases}
\end{figure}

We analyze failure cases to understand the limitations of our method. The main failure modes we observe include: (1) \textbf{Text generation distortion}: Images containing text or fine typography may lose readability after stylization, as the style transfer process can alter character shapes and stroke patterns, making text illegible. (2) \textbf{Complex hand gestures}: Complex hand poses and gestures may exhibit distortion or unnatural deformation during style transfer, particularly when the style requires significant geometric transformation. These cases are particularly challenging because text and hand structures require precise geometric preservation at fine-grained levels. Representative failure cases illustrating these limitations are shown in Fig.~\ref{fig:failure_cases}, where we compare UniCSG with Nano-banana across six different styles in both text-guided and reference-guided settings.

\textbf{Text distortion case}: In terms of text preservation, UniCSG performs slightly worse than Nano-banana. However, UniCSG achieves superior overall stylization quality, particularly in Ghibli and Paper cutting styles. Nano-banana tends to overfit to styles, leading to uncontrolled generation. Overall, reference-guided results outperform text-guided results due to the additional style constraints provided by reference images.

\textbf{Hand gesture deformation case}: In text-guided tasks, UniCSG successfully preserves hand structures in clay and Paper cutting styles. Nano-banana succeeds in text-guided tasks except for 3D deformation styles where it fails. In reference-guided tasks, Nano-banana successfully maintains hand structures in clay style. Generally, text-guided tasks preserve content more easily, while deformation styles pose greater challenges. Nano-banana tends to stylize on fully realistic content, which easily achieves excessive content preservation but insufficient stylization effects.

These failure cases reveal that UniCSG still needs further improvement in preserving fine-grained content details such as text readability and complex hand structures, especially when balancing between content preservation and stylization quality. Future work will focus on enhancing content preservation capabilities while maintaining strong stylization performance.

\subsection{More Qualitative Results}

\begin{figure*}
\centering
\includegraphics[width=0.9\textwidth]{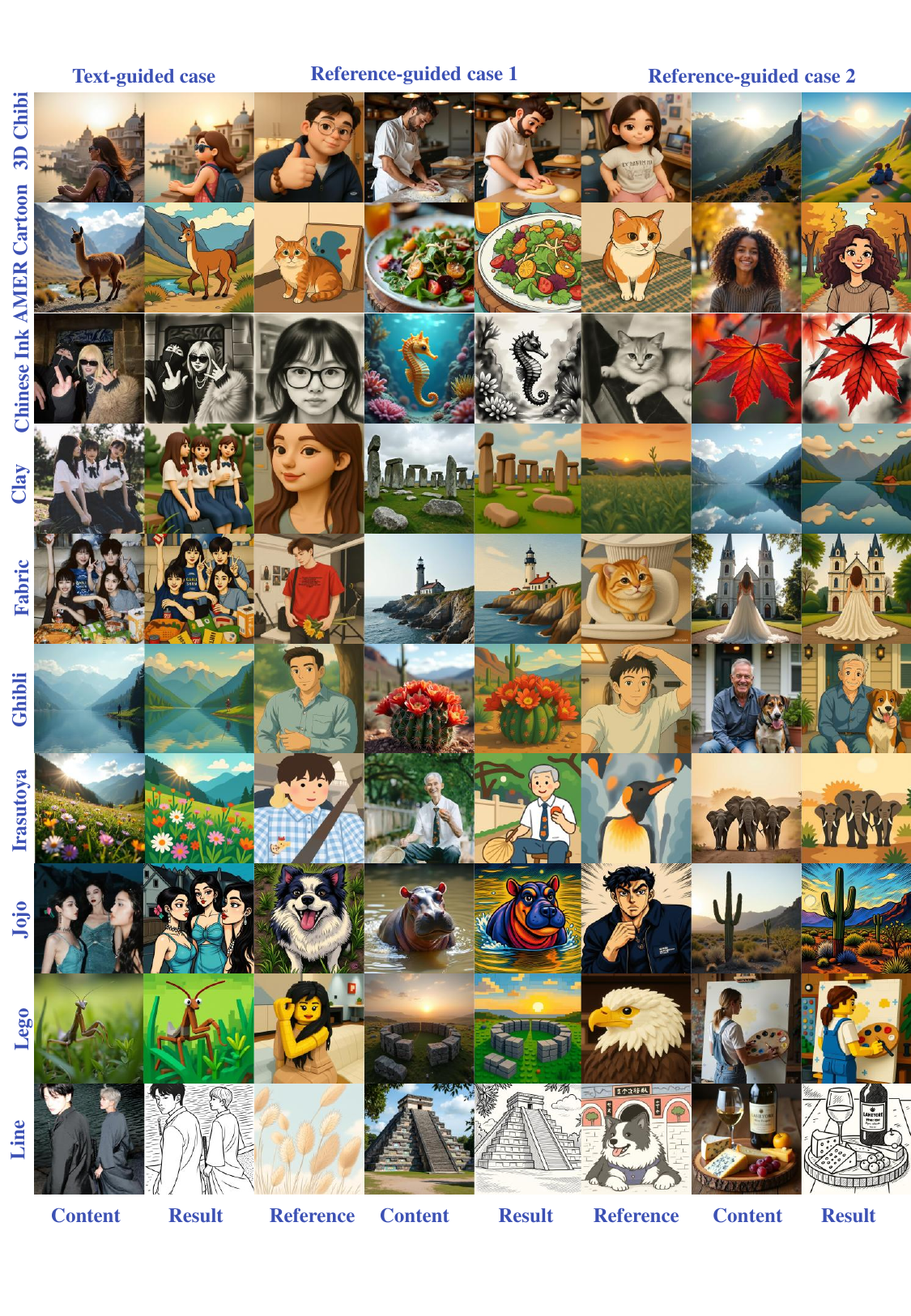}
\caption{Extended qualitative results showcasing UniCSG's performance across diverse content types and style categories. The figure includes various cases demonstrating the method's effectiveness on different content domains and style categories.}
\label{fig:extended_qualitative1}
\end{figure*}

\begin{figure*}
\centering
\includegraphics[width=0.9\textwidth]{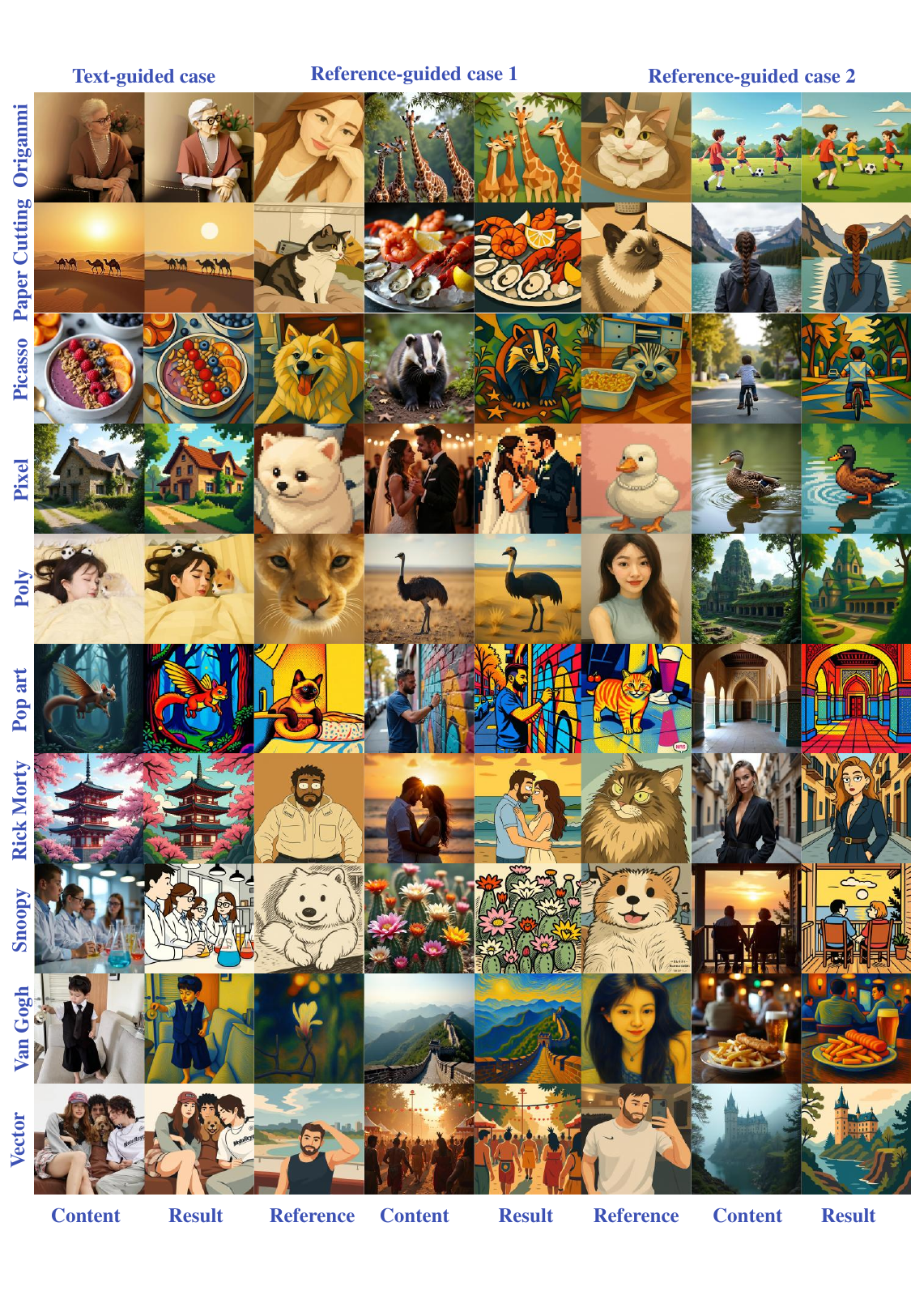}
\caption{Extended qualitative results showcasing UniCSG's performance across diverse content types and style categories(continued).}
\label{fig:extended_qualitative2}
\end{figure*}

We provide additional qualitative results showcasing UniCSG's performance across diverse content types and style categories. Fig.~\ref{fig:extended_qualitative1} and Fig.~\ref{fig:extended_qualitative2} presents various cases demonstrating the method's effectiveness, including examples from different content domains (portraits, landscapes, architecture, animals, objects, food) and diverse style categories. These examples further validate the robustness and generalization capability of our method across different scenarios.

\end{document}